\PassOptionsToPackage{table}{xcolor}
\documentclass{article}

\usepackage{arxiv}

\usepackage[utf8]{inputenc}
\usepackage[T1]{fontenc}
\usepackage{hyperref}
\usepackage{url}
\usepackage{xcolor}
\usepackage{booktabs}
\usepackage{amsfonts}
\usepackage{amsmath}
\usepackage{cleveref}
\usepackage{amssymb}
\usepackage{nicefrac}
\usepackage{microtype}
\usepackage{graphicx}
\usepackage{array}
\usepackage{flafter}
\usepackage{placeins}
\usepackage{caption}
\usepackage{rotating}
\usepackage{float}
\usepackage[numbers,square]{natbib}
\usepackage{doi}
\usepackage{comment}

\graphicspath{{../report/}{../report/figures/}}

\newcommand{\vcentered}[1]{\raisebox{-0.5\height}{#1}}
\newcommand{\rowlab}[1]{\makebox[1.2em][c]{\vcentered{\rotatebox[origin=c]{90}{\makebox[0.11\textwidth][c]{\small #1}}}}}
\newcommand{\collab}[1]{\small #1}
\newcommand{\scell}[1]{\vcentered{\includegraphics[width=0.232\linewidth]{figures/sd35_preview/sd35_#1}}}
\newcommand{\fcell}[1]{\vcentered{\includegraphics[width=0.232\linewidth]{figures/flux_preview/flux_#1}}}
\newcommand{\pcell}[1]{\vcentered{\includegraphics[width=0.152\textwidth]{figures/pic3/#1}}}
\newcommand{\cfgcollab}[1]{\makebox[0.115\textwidth][c]{\normalsize #1}}
\newcommand{\cfgrowlab}[1]{\makebox[2.8em][c]{\vcentered{\rotatebox[origin=c]{90}{\makebox[0.115\textwidth][c]{\normalsize #1}}}}}
\newcommand{\cfgspacer}{\makebox[2.8em][c]{}}
\newcommand{\cfgcell}[1]{\vcentered{\includegraphics[width=0.115\textwidth]{figures/cfgsweep/#1}}}

\definecolor{resultgray}{RGB}{238,238,238}
\definecolor{resultgreen}{RGB}{222,242,226}
\definecolor{resultgreenstrong}{RGB}{190,229,198}
\definecolor{resultred}{RGB}{250,226,228}
\definecolor{resultredstrong}{RGB}{245,198,202}
\newcommand{\resulttie}[1]{\cellcolor{resultgray}#1}
\newcommand{\resultgain}[1]{\cellcolor{resultgreen}#1}
\newcommand{\resultgainstrong}[1]{\cellcolor{resultgreenstrong}#1}
\newcommand{\resultloss}[1]{\cellcolor{resultred}#1}
\newcommand{\resultlossstrong}[1]{\cellcolor{resultredstrong}#1}

\title{Revisiting Classifier-Free Guidance Methods in Latent Diffusion Models}

\author{%
    Artem Sergievskii\thanks{Equal contribution.} \\
    HSE University
    \And
    Artyom Turevich\footnotemark[1] \\
    HSE University \\
    \And
    Sergey Kastryulin \\ 
    Yandex, HSE University \\
  }

\renewcommand{\shorttitle}{Revisiting Classifier-Free Guidance Methods}

\begin{document}

\maketitle

\begin{abstract}
Inference-time quality-enhancement methods are an effective and widely adopted
means of improving diffusion models without expensive retraining. We study a
family of training-free techniques conceptually rooted in Classifier-Free
Guidance (CFG), most of which were originally proposed on older U-Net diffusion
models and validated using metrics that assess image quality in isolation, without accounting for compositional alignment or semantic correspondence between the generated image and its associated text prompt. We re-evaluate eight such
methods on two open-weight rectified-flow transformers under a fixed per-model
protocol and three compositional-alignment benchmarks. No method consistently
improves on CFG across the measured criteria. APG obtains several nominal best scores, but the corresponding gains often remain within the estimated evaluation uncertainty.
Attention-perturbation methods provide isolated gains on SD3.5 Medium and more frequent
degradations on FLUX.2~[klein] 4B Base, while CFG remains a competitive lower-cost baseline.
\end{abstract}

\section{Introduction}
\label{sec:intro}

Classifier-free guidance (CFG)~\cite{ho2022cfg} is the standard inference-time
control in conditional diffusion models. It improves prompt alignment without
changing model weights, but high guidance scales can reduce diversity and
introduce oversaturation or structural artifacts. These failures motivated
several training-free alternatives.

The alternatives are difficult to compare because their studies use different
models, datasets, resolutions, and they were published at different times, often sequentially. Most evidence comes from U-Net
Stable Diffusion variants or class-conditional ImageNet models and emphasizes
FID, CLIP score, or Inception Score. These metrics do not directly test
counting, relations, dense prompt following, or text rendering. 

We compare eight methods under a fixed per-model protocol on two models:
Stable Diffusion~3.5 Medium~\cite{esser2024sd3,sd35medium} and
FLUX.2~[klein] 4B Base~\cite{flux2klein}. Evaluation uses GenEval~\cite{ghosh2023geneval},
DPG-Bench~\cite{hu2024ella}, three selected OneIG-Bench
families~\cite{chang2026oneig}, and paired prompt bootstrap. Only a few gains exceed
the shared reference margins: SAG improves selected SD3.5 metrics, while no alternative
does so on the FLUX headline metrics.

The comparison therefore does not identify a general replacement for CFG.
Nominal leaders change across benchmarks. The alternatives are better treated
as model- and metric-specific choices than as universal upgrades.

We contribute implementations and configurations for both
models,\footnote{\url{https://github.com/ThereWillComeSoftRains/RevisitingCFGMethods}}
a controlled comparison with benchmark-independent parameter selection, and an
uncertainty-aware evaluation with qualitative examples and a comparison with the metrics and gains reported in the original papers.

\section{Related Work}
\label{sec:related}

\paragraph{Guidance for diffusion models.}
Classifier guidance uses an external classifier to trade diversity for sample
fidelity~\cite{dhariwal2021adm}. CFG removes this auxiliary network by jointly
learning conditional and unconditional predictions~\cite{ho2022cfg}. Its scale
controls prompt adherence, but large values can amplify saturation and
artifacts. This trade-off makes inference-time guidance useful without making
its scale or update rule interchangeable across models.

The methods studied here follow two broad designs. CFG++~\cite{chung2024cfgpp},
CFG-Zero*~\cite{fan2025cfgzero}, APG~\cite{sadat2025apg}, and
TCFG~\cite{kwon2025tcfg} modify the CFG update through interpolation,
initialization, projection, or damping. SAG~\cite{hong2023sag},
PAG~\cite{ahn2024pag}, SEG~\cite{hong2024seg}, and OSEG~\cite{fahim2026oseg}
instead construct a weaker prediction by perturbing attention and guide away
from it. Related training-free approaches change the time interval over which
guidance is applied, anneal the condition, or use a weak model, as in interval
guidance~\cite{kynkaanniemi2024interval}, CADS~\cite{sadat2024cads}, and
autoguidance~\cite{karras2024autoguidance}. They fall outside our controlled
set of CFG-derived updates.

\paragraph{Evaluation of text-to-image models.}
Original guidance papers commonly report global quality or similarity metrics
under different model, sampler, and dataset choices. This demonstrates gains in
their intended settings, but does not provide a direct ranking on current
text-to-image systems.
Newer benchmarks test prompt-level behavior: TIFA uses visual
questions~\cite{hu2023tifa}, T2I-CompBench measures compositional
relations~\cite{huang2023t2icompbench}, GenEval focuses on object composition,
and DPG-Bench evaluates dense prompts. OneIG-Bench separates object, text,
reasoning, and style abilities, while ImageReward~\cite{xu2023imagereward}
models human preferences. These metrics are complementary rather than
interchangeable.

\paragraph{Transfer to transformer backbones.}
Most studied methods were developed or first tested on U-Net architectures.
Prediction-level modifications require few structural choices, but attention
perturbation depends on the modified blocks. Diffusion transformers use joint,
dual-stream, or single-stream layouts without direct U-Net counterparts, and
perturbation guidance is sensitive to layer and head
selection~\cite{ahn2025headhunter}. We therefore include adaptation and layer
selection in the comparison rather than reusing the original U-Net choices
without validation.

\section{Setup and Protocol}
\label{sec:method}

\paragraph{Benchmark scope.}
We compare no-CFG and vanilla CFG with CFG++, CFG-Zero*, APG, TCFG, SAG, PAG,
SEG, and OSEG. Each method is an inference-time modification of a frozen model,
with its definition fixed within each model. Sampling and evaluation settings
are held constant so that only the guidance rule and its selected parameters
change.

\paragraph{Model adaptation.} Prediction-level methods are matched to each
model's parameterization and scheduler. Attention methods additionally require
transformer block selection, performed separately per model. We denote the
$N$-th dual text--image block by \texttt{dN} and the $N$-th self-only block by
\texttt{sN}. This notation keeps the reported choices comparable despite the
different transformer layouts.

\paragraph{Hyperparameter selection.}
Parameters are fixed before benchmark evaluation. Starting from the ranges in
each paper, we use a small fixed prompt set and vary one parameter at a time:
guidance and method-specific scales, ranks, blur strengths, or layer choices.
The prompts cover simple scenes, multiple objects, and spatial relations.
We reject settings that repeatedly produce oversaturation, halos, duplicated
structures, loss of detail, or collapse, then choose a prompt-faithful candidate
without a consistent visible failure. No benchmark score is used for selection.
This avoids tuning directly to the reported test sets while keeping the search
small enough to repeat for both models.
Final settings and representative sweep grids are provided in
Appendices~\hyperref[sec:app-hparams]{\ref*{sec:app-hparams}}
and~\hyperref[sec:app-sweeps]{\ref*{sec:app-sweeps}}.

\paragraph{Models and evaluation.} We evaluate on two open-weight rectified-flow
models: Stable Diffusion~3.5 Medium~\cite{esser2024sd3,sd35medium}, a 2.5B MM-DiT
with joint text--image attention, and FLUX.2~[klein] 4B Base~\cite{flux2klein}, a 4B
hybrid dual/single-stream transformer. We use the full prompt sets on selected benchmarks. Methods share prompts, seeds, scheduler,
resolution, and sampling budget within each model: 25 steps for SD3.5 and 30 for
FLUX; inference used V100 32GB GPUs. GenEval measures object composition,
DPG-Bench dense-prompt faithfulness, and the selected OneIG families cover
General Object, Text Rendering, and Knowledge Reasoning. These tasks probe
different alignment failures rather than a single aggregate quality dimension.
Style-oriented OneIG families were not generated.

\paragraph{Uncertainty.}
We use 10,000 paired prompt-bootstrap replicates with synchronized sampling for
every method and CFG. Each replicate samples the original number of prompts with
replacement and applies the same indices to all methods. GenEval is resampled
within tasks, and the nonlinear OneIG text score is recomputed per replicate.
For each benchmark, centered method--CFG errors are pooled over guided
alternatives; their 95th percentile is the shared bootstrap margin. Differences inside it
are treated as unresolved rather than as evidence of equivalence.

\section{Results}
\label{sec:results}

\subsection{Quantitative Results}
\label{sec:results-quant}

\Cref{tab:summary-both} reports the five headline scores for both models. The
same shared paired-bootstrap margins are shown above each model block. Colors
encode the difference from CFG under this rule, while bold identifies the
nominal column maximum independently of the margin.

\begin{table}[!htb]\centering\scriptsize
\setlength{\tabcolsep}{2.0pt}
\renewcommand{\arraystretch}{1.0}
\caption{Headline scores for both models. GE is GenEval; DPG is mean$\times$100;
Obj, Txt, and Rsn are the selected OneIG families. Gray denotes differences
within one shared reference margin of CFG; green/red denote larger gains/losses. A
stronger shade beyond two margins is used only for visual emphasis. Bold marks the column maximum.}
\label{tab:summary-both}
\resizebox{\textwidth}{!}{%
\begin{tabular}{@{}l*{5}{c}@{\hspace{3pt}{\color{black!30}\vrule width 0.35pt}\hspace{3pt}}*{5}{c}@{}}
\toprule
& \multicolumn{5}{c}{\textbf{SD3.5 Medium}}
& \multicolumn{5}{c}{\textbf{FLUX.2~[klein] 4B Base}} \\
\cmidrule(lr){2-6}\cmidrule(lr){7-11}
Method & GE & DPG & Obj & Txt & Rsn & GE & DPG & Obj & Txt & Rsn \\
95\% $\Delta$ margin & $\pm$0.027 & $\pm$0.76 & $\pm$0.019 & $\pm$0.044 & $\pm$0.005 & $\pm$0.028 & $\pm$0.85 & $\pm$0.018 & $\pm$0.042 & $\pm$0.004 \\
\midrule
no-CFG & \resultlossstrong{0.367} & \resultlossstrong{73.51} & \resultlossstrong{0.526} & \resultlossstrong{0.037} & \resultlossstrong{0.160} & \resultlossstrong{0.346} & \resultlossstrong{73.22} & \resultlossstrong{0.666} & \resultlossstrong{0.177} & \resultlossstrong{0.188} \\
CFG & \resulttie{0.655} & \resulttie{84.35} & \resulttie{0.713} & \resulttie{0.443} & \resulttie{0.257} & \resulttie{0.786} & \resulttie{83.30} & \resulttie{0.782} & \resulttie{0.736} & \resulttie{0.281} \\
\midrule
CFG++ & \resultloss{0.617} & \resulttie{84.29} & \resultgainstrong{0.749} & \resulttie{0.459} & \resultlossstrong{0.232} & \resultlossstrong{0.704} & \resulttie{83.35} & \resulttie{0.784} & \resulttie{0.753} & \resulttie{\textbf{0.284}} \\
CFG-Zero* & \resulttie{0.638} & \resulttie{84.19} & \resulttie{0.720} & \resulttie{0.409} & \resultlossstrong{0.231} & \resultlossstrong{0.726} & \resulttie{82.91} & \resulttie{\textbf{0.792}} & \resulttie{0.723} & \resulttie{0.279} \\
APG & \resulttie{0.671} & \resulttie{84.54} & \resultgainstrong{\textbf{0.763}} & \resulttie{\textbf{0.471}} & \resultloss{0.250} & \resulttie{\textbf{0.792}} & \resulttie{\textbf{83.50}} & \resulttie{0.781} & \resulttie{\textbf{0.768}} & \resulttie{0.283} \\
TCFG & \resulttie{0.644} & \resultloss{83.45} & \resultgain{0.739} & \resulttie{0.426} & \resultlossstrong{0.245} & \resulttie{0.774} & \resulttie{83.00} & \resulttie{0.776} & \resulttie{0.734} & \resulttie{0.278} \\
SAG & \resultgainstrong{\textbf{0.715}} & \resulttie{84.41} & \resulttie{0.725} & \resulttie{0.408} & \resultgainstrong{\textbf{0.269}} & \resulttie{0.762} & \resulttie{83.33} & \resulttie{0.776} & \resultloss{0.681} & \resultlossstrong{0.273} \\
SEG & \resultloss{0.623} & \resulttie{84.03} & \resulttie{0.697} & \resulttie{0.434} & \resultlossstrong{0.243} & \resultlossstrong{0.724} & \resulttie{83.21} & \resulttie{0.785} & \resultlossstrong{0.648} & \resultlossstrong{0.255} \\
OSEG & \resulttie{0.650} & \resulttie{\textbf{84.61}} & \resulttie{0.723} & \resultloss{0.384} & \resulttie{0.259} & \resultloss{0.743} & \resulttie{83.11} & \resulttie{0.772} & \resultloss{0.662} & \resultlossstrong{0.261} \\
PAG & \resultlossstrong{0.595} & \resultlossstrong{82.56} & \resulttie{0.702} & \resultlossstrong{0.351} & \resultlossstrong{0.220} & \resultloss{0.735} & \resulttie{83.35} & \resultloss{0.763} & \resultloss{0.694} & \resultlossstrong{0.267} \\
\bottomrule
\end{tabular}%
}
\end{table}

\FloatBarrier

The no-CFG baseline falls below CFG by more than two margins on every headline
metric for both models. This shows that guidance is beneficial under the tested
sampling regimes, but the alternatives to CFG show a less uniform pattern.
Only a small number of their nominal gains exceed the shared margins, whereas
resolved decreases occur across several methods and metrics.

On SD3.5, SAG produces the largest GenEval score and improves reasoning beyond
the margin. CFG++, APG, and TCFG exceed the margin on OneIG object, with APG
obtaining the nominal maximum. These gains remain local to particular metrics.
CFG++ and CFG-Zero* decrease reasoning, CFG++ also decreases GenEval, and APG
and TCFG have resolved losses on reasoning. Among attention-based methods, SEG
decreases GenEval and reasoning, OSEG decreases text rendering, and PAG has
resolved losses on GenEval, DPG-Bench, text rendering, and reasoning. No
positive DPG-Bench difference exceeds its margin.

The FLUX.2~[klein] 4B Base comparison contains no positive change beyond a shared margin on any
headline metric. APG nominally leads GenEval, DPG-Bench, and text rendering, but
all three differences from CFG remain unresolved. CFG++ and CFG-Zero*  have GenEval losses beyond margin. SEG, OSEG, and
PAG also decrease GenEval, and each attention-based method decreases text
rendering, reasoning, or both. The SD3.5 gains of SAG therefore do not transfer
to FLUX under the selected configuration.

\hyperref[sec:app-details]{Appendix~\ref*{sec:app-details}} gives the GenEval
category scores and DPG-Bench L1/L2 breakdowns. These tables help identify which
prompt properties contribute to an aggregate result, but their isolated maxima
should not be interpreted as a general method ranking. All evaluated OneIG
families already appear in \cref{tab:summary-both}.

\subsection{Qualitative Overview}
\label{sec:results-qual}

\begin{figure}[!htbp]\centering
\setlength{\tabcolsep}{0pt}\renewcommand{\arraystretch}{0.6}
\begin{minipage}[t]{0.499\textwidth}\centering
{\small SD3.5 Medium}\\[1pt]
\begin{tabular}{@{}c@{}cccc@{}}
 & \collab{portrait} & \collab{dog/tie} & \collab{valley} & \collab{horse} \\
\rowlab{no-CFG}    & \scell{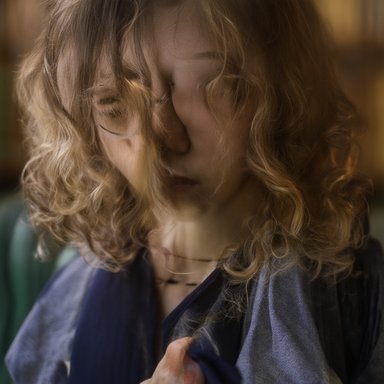}    & \scell{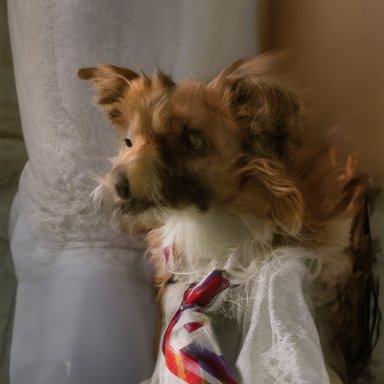}    & \scell{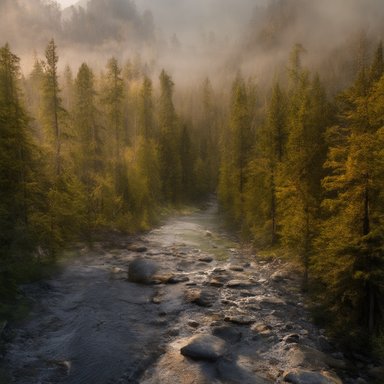}    & \scell{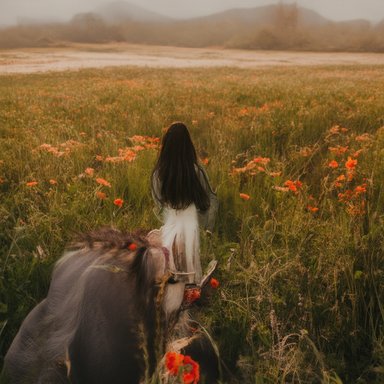} \\
\rowlab{CFG}       & \scell{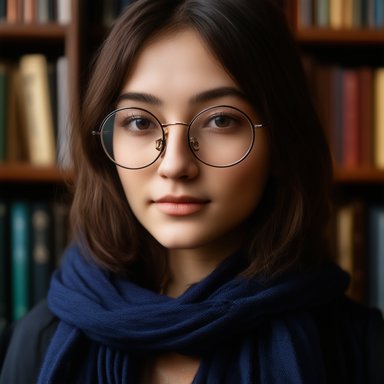}       & \scell{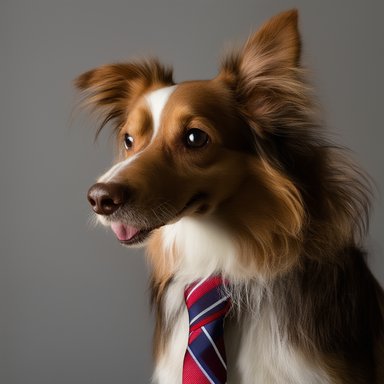}       & \scell{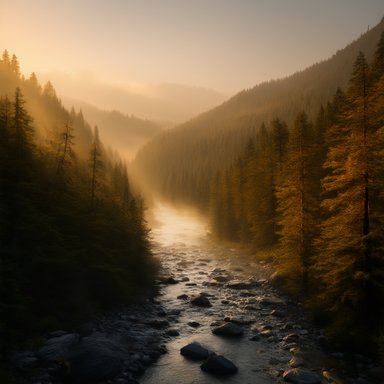}       & \scell{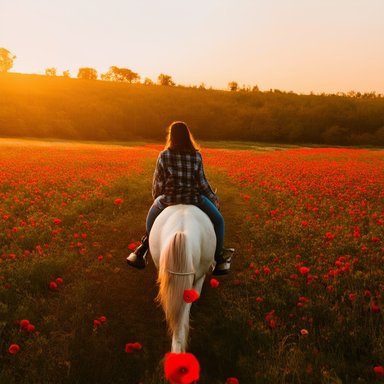} \\
\rowlab{CFG++}     & \scell{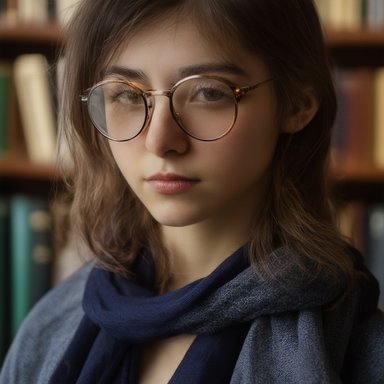}     & \scell{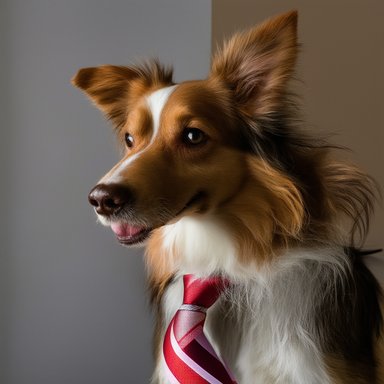}     & \scell{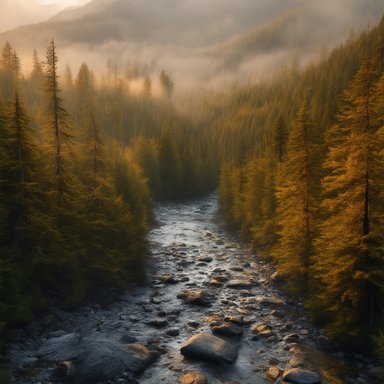}     & \scell{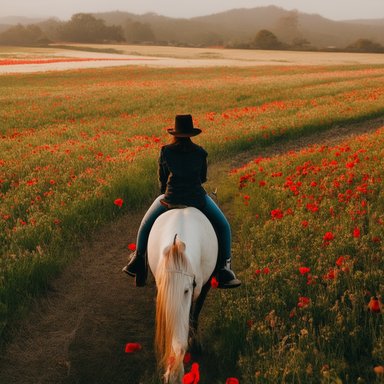} \\
\rowlab{CFG-Zero*} & \scell{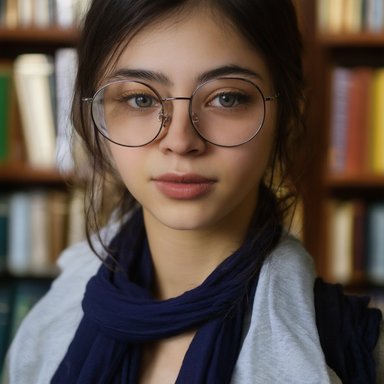}     & \scell{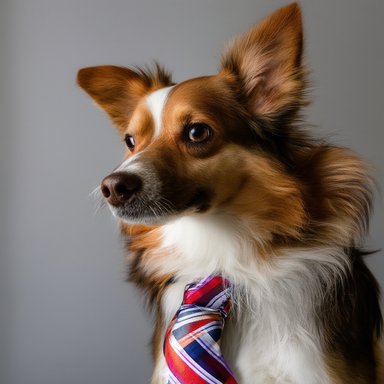}     & \scell{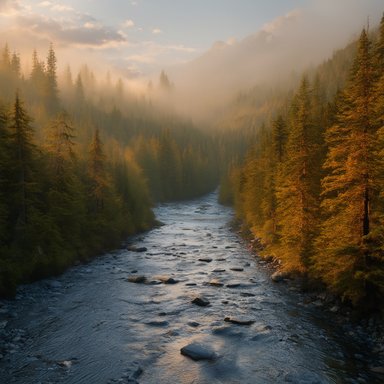}     & \scell{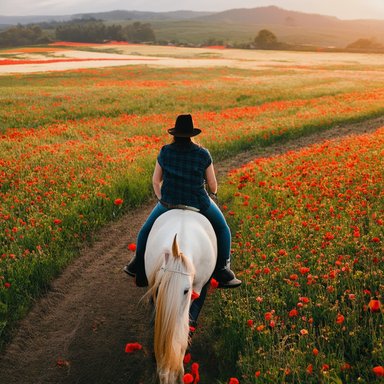} \\
\rowlab{APG}       & \scell{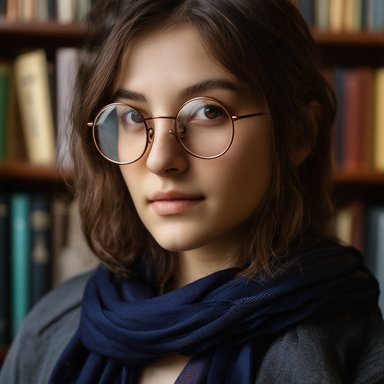}       & \scell{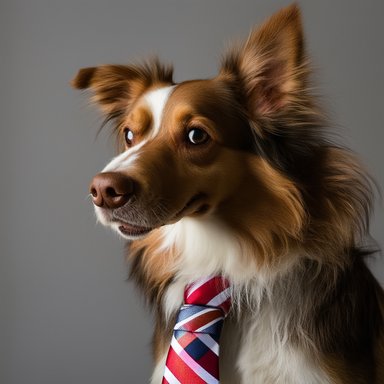}       & \scell{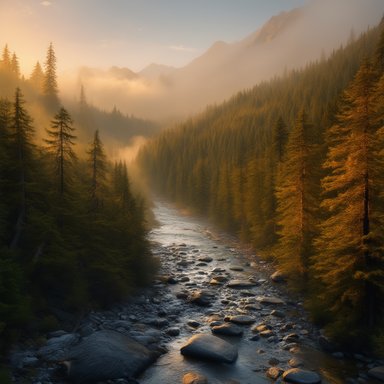}       & \scell{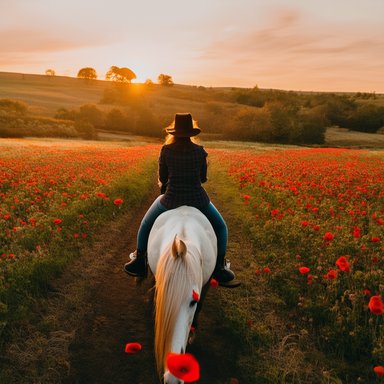} \\
\rowlab{TCFG}      & \scell{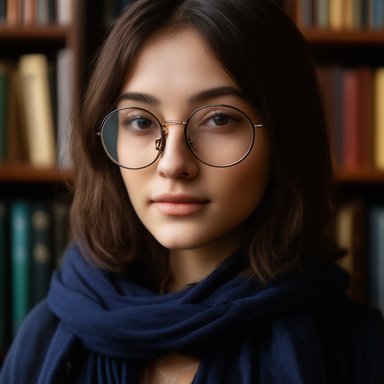}      & \scell{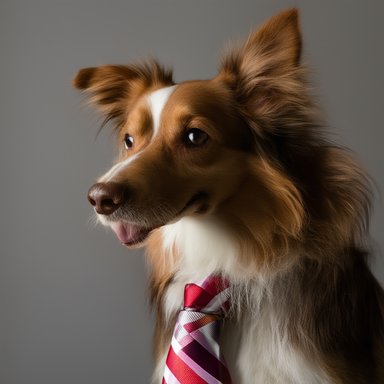}      & \scell{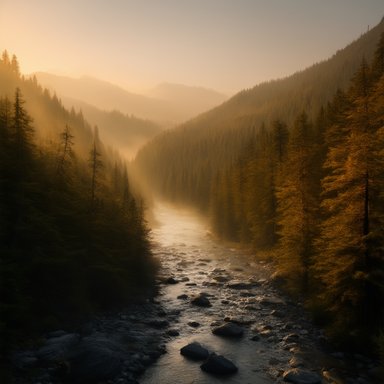}      & \scell{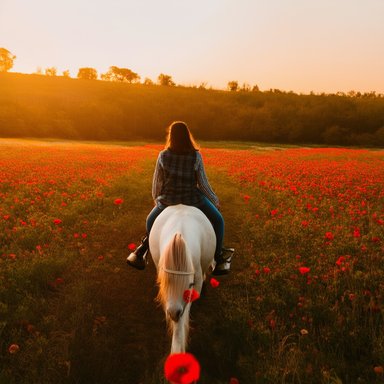} \\
\rowlab{SAG}       & \scell{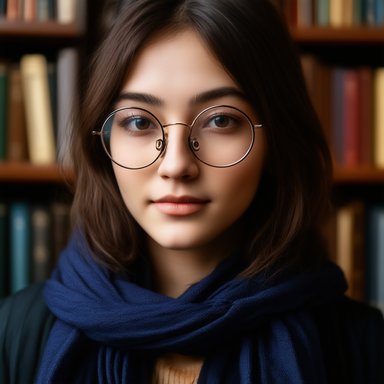}       & \scell{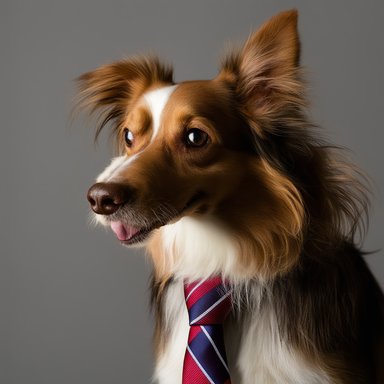}       & \scell{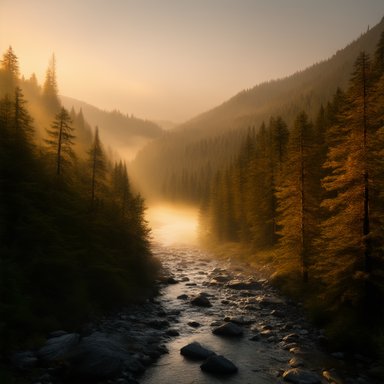}       & \scell{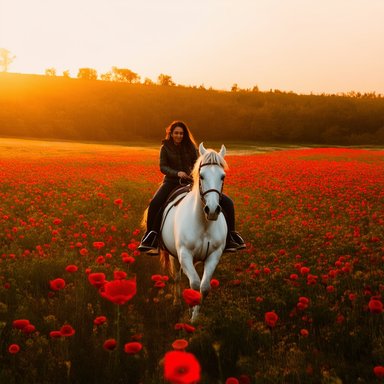} \\
\rowlab{OSEG}      & \scell{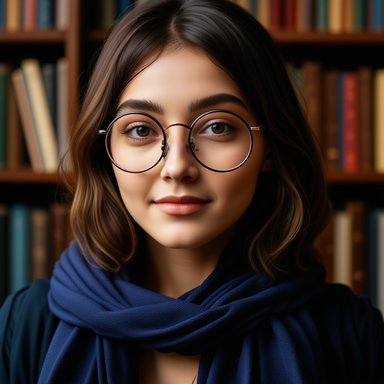}      & \scell{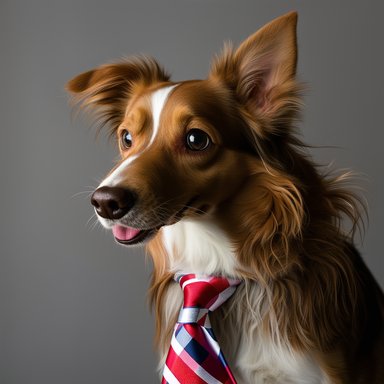}      & \scell{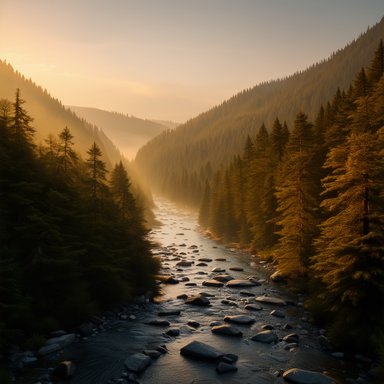}      & \scell{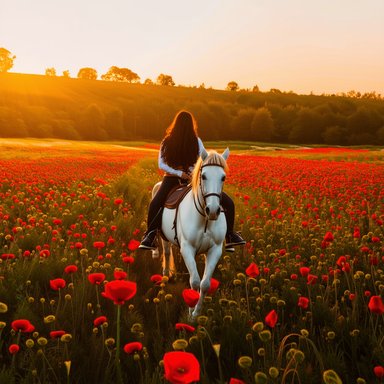} \\
\rowlab{PAG}       & \scell{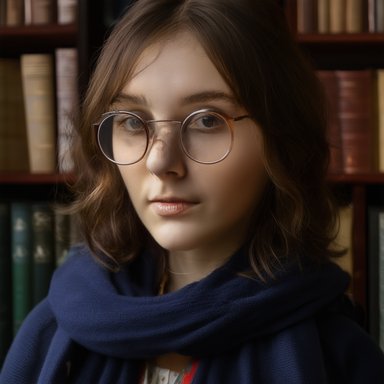}       & \scell{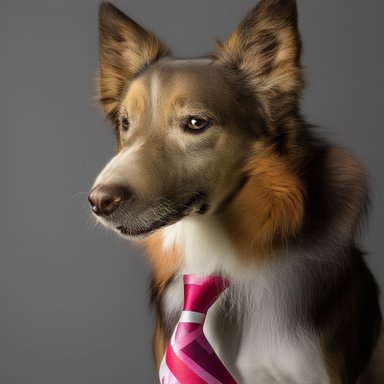}       & \scell{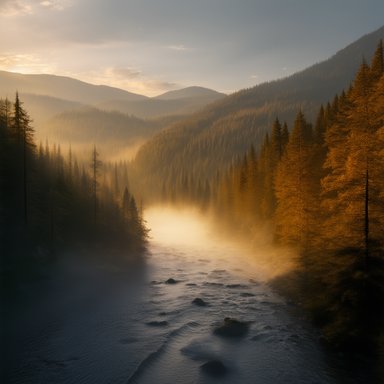}       & \scell{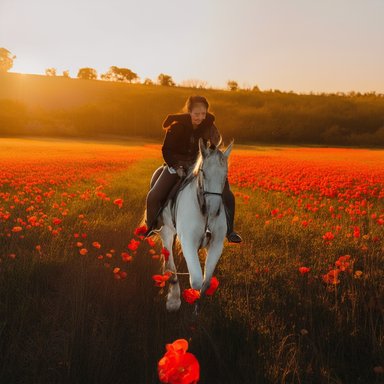} \\
\rowlab{SEG}       & \scell{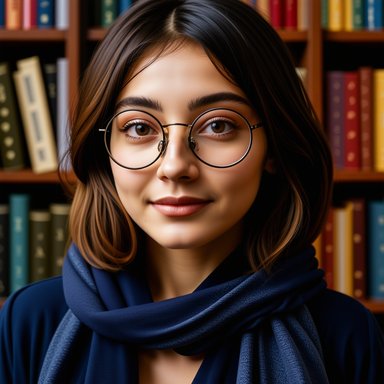} & \scell{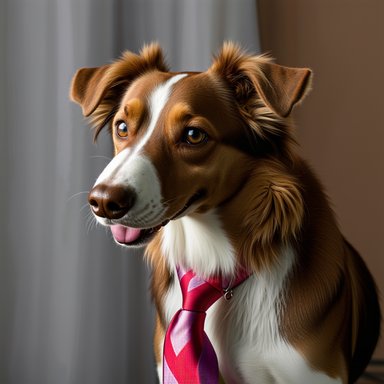} & \scell{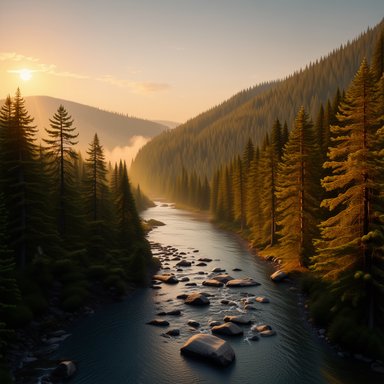} & \scell{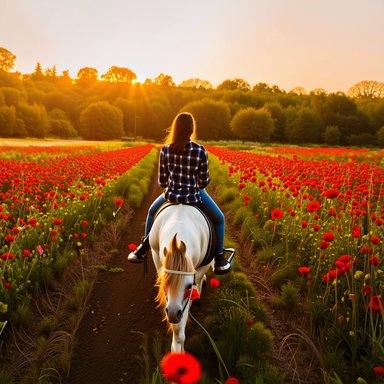} \\
\end{tabular}
\end{minipage}\hfill
\begin{minipage}[t]{0.499\textwidth}\centering
{\small FLUX.2~[klein] 4B Base}\\[1pt]
\begin{tabular}{@{}c@{}cccc@{}}
 & \collab{portrait} & \collab{dog/tie} & \collab{valley} & \collab{horse} \\
\rowlab{no-CFG}    & \fcell{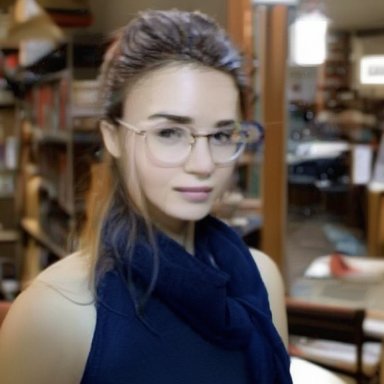}    & \fcell{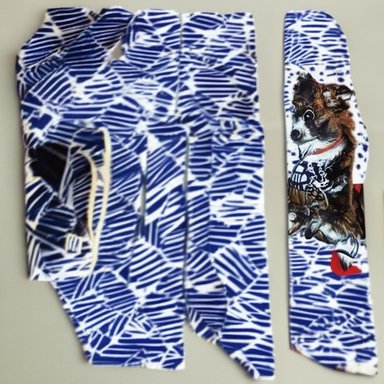}    & \fcell{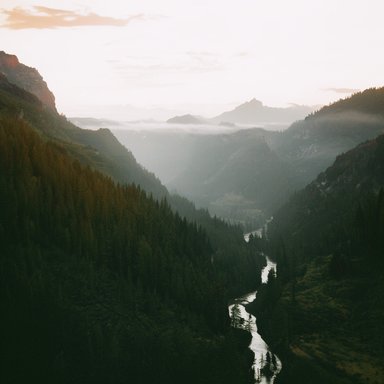}    & \fcell{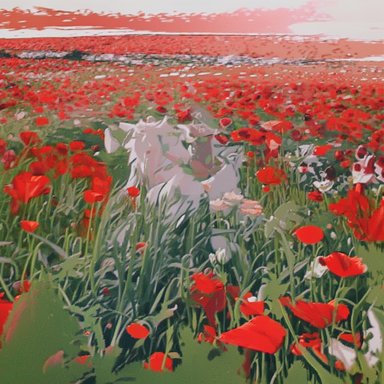} \\
\rowlab{CFG}       & \fcell{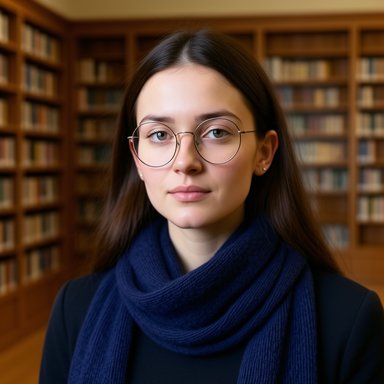}       & \fcell{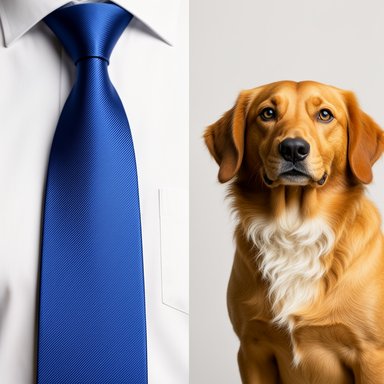}       & \fcell{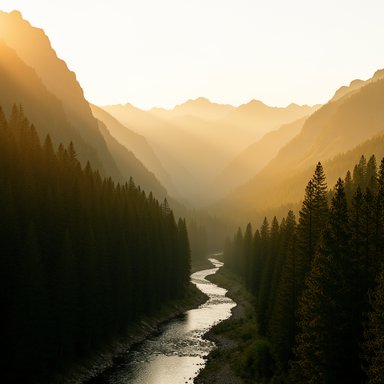}       & \fcell{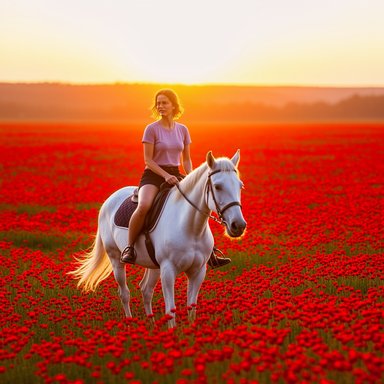} \\
\rowlab{CFG++}     & \fcell{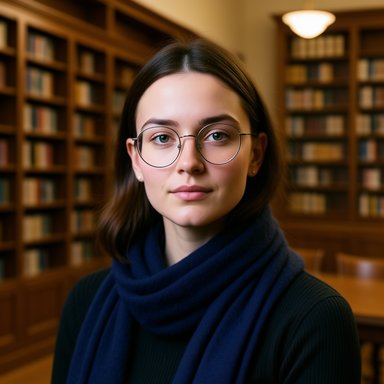}     & \fcell{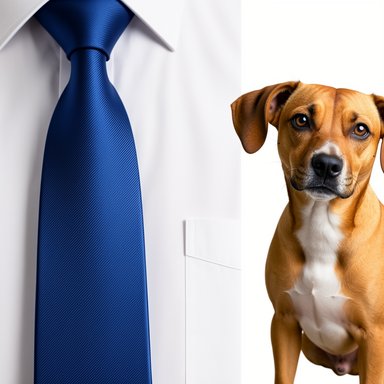}     & \fcell{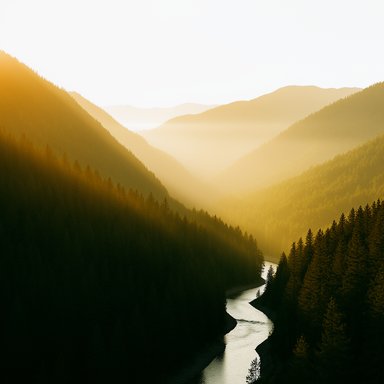}     & \fcell{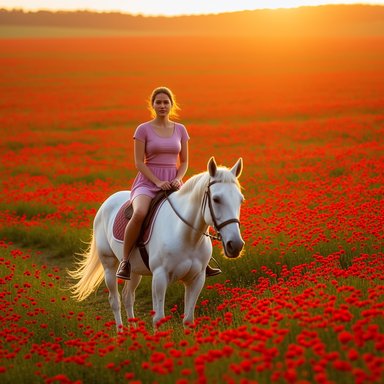} \\
\rowlab{CFG-Zero*} & \fcell{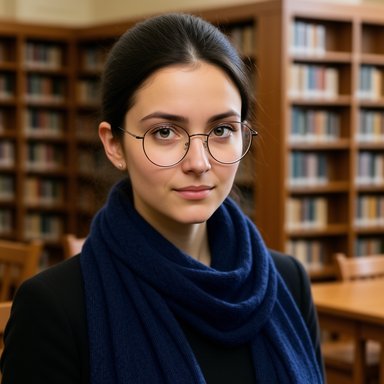}     & \fcell{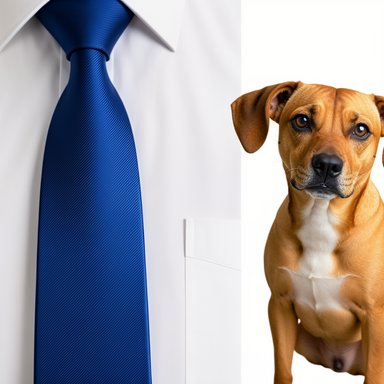}     & \fcell{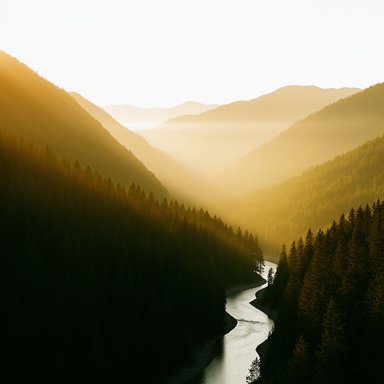}     & \fcell{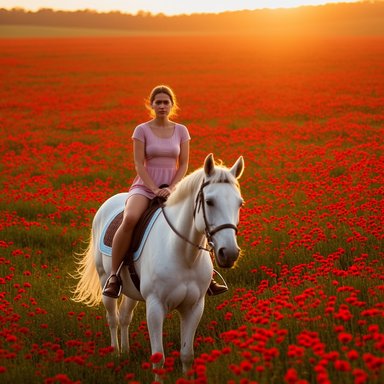} \\
\rowlab{APG}       & \fcell{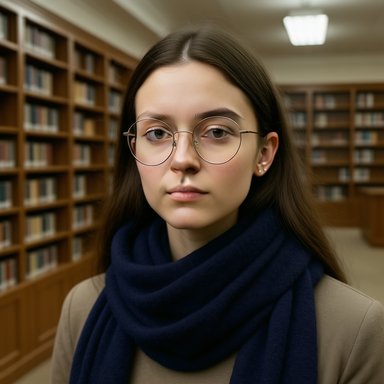}       & \fcell{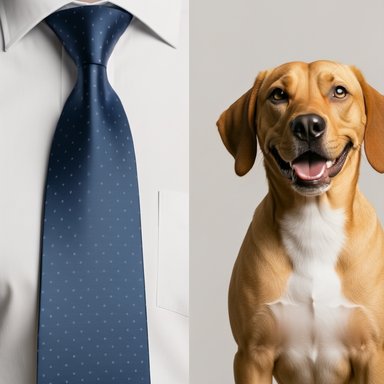}       & \fcell{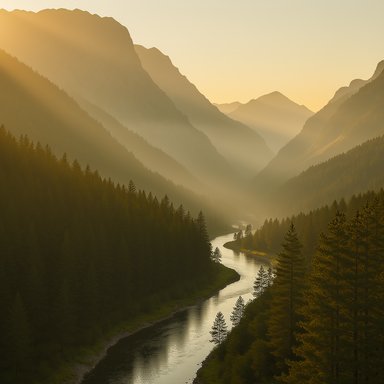}       & \fcell{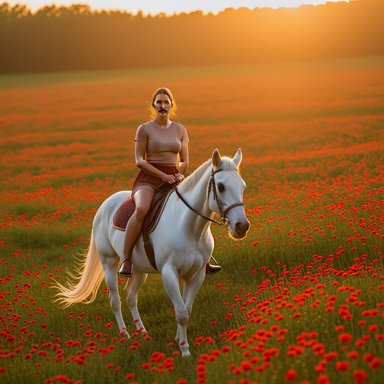} \\
\rowlab{TCFG}      & \fcell{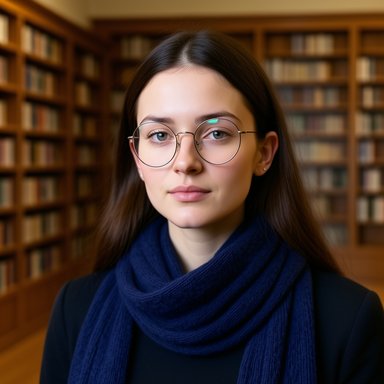}      & \fcell{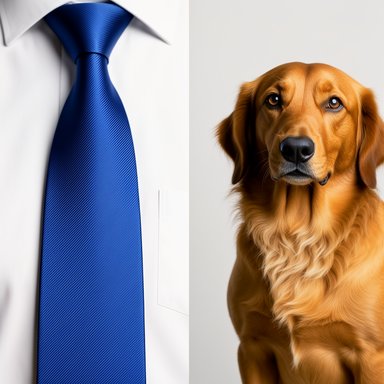}      & \fcell{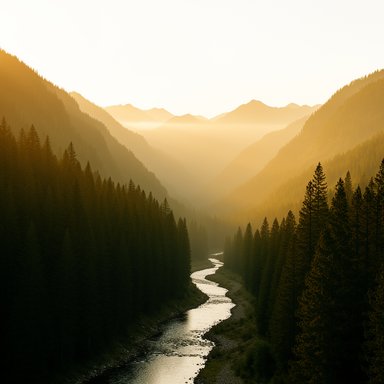}      & \fcell{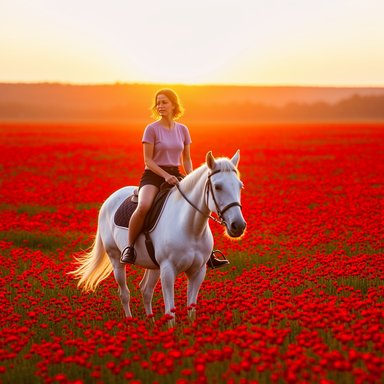} \\
\rowlab{SAG}       & \fcell{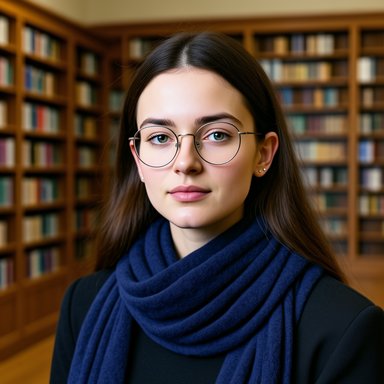}       & \fcell{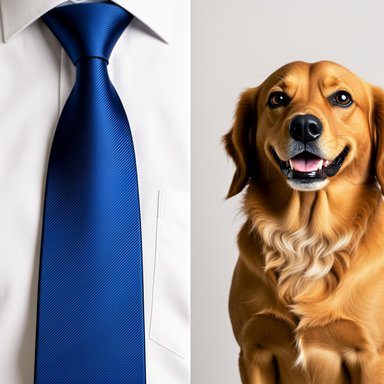}       & \fcell{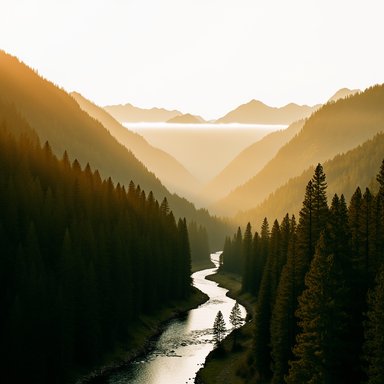}       & \fcell{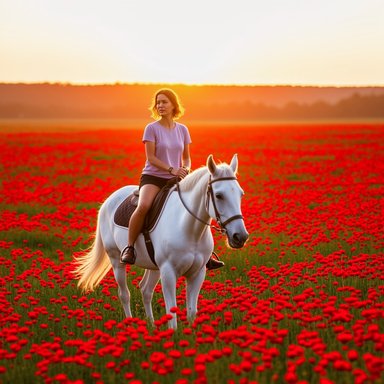} \\
\rowlab{OSEG}      & \fcell{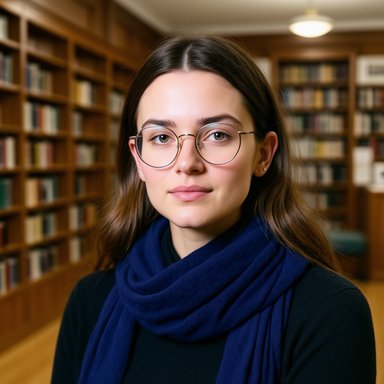}      & \fcell{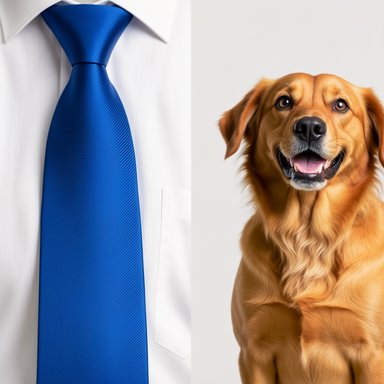}      & \fcell{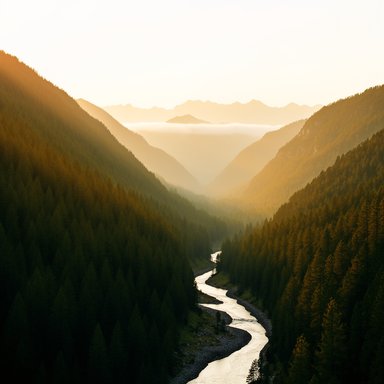}      & \fcell{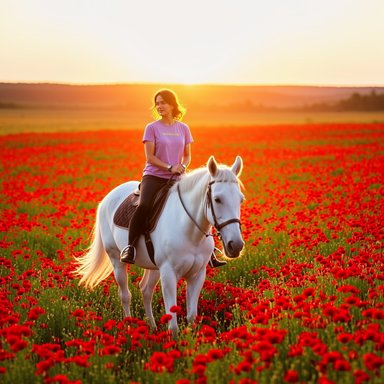} \\
\rowlab{PAG}       & \fcell{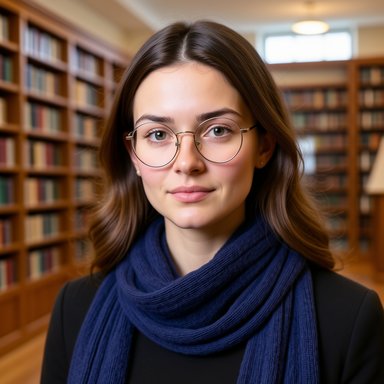}       & \fcell{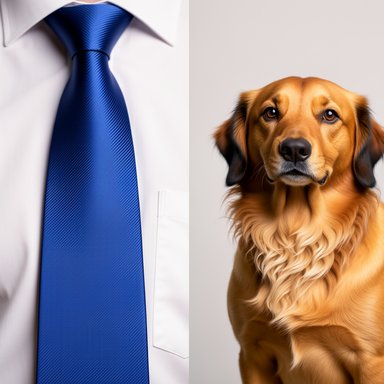}       & \fcell{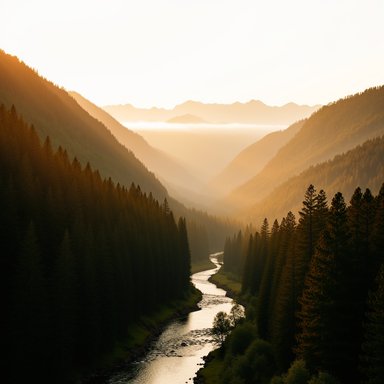}       & \fcell{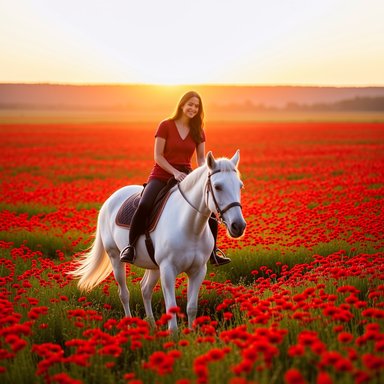} \\
\rowlab{SEG}       & \fcell{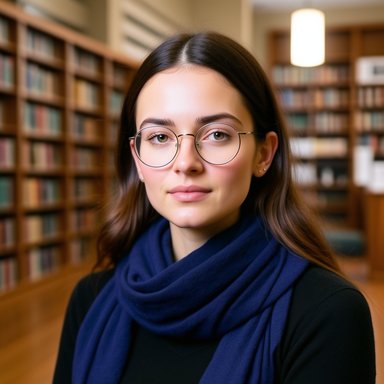} & \fcell{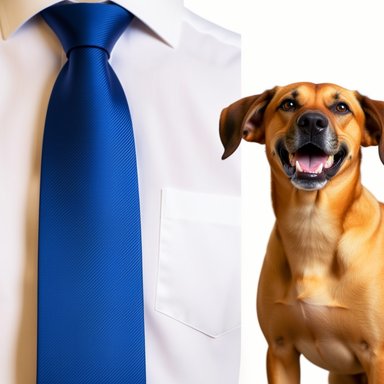} & \fcell{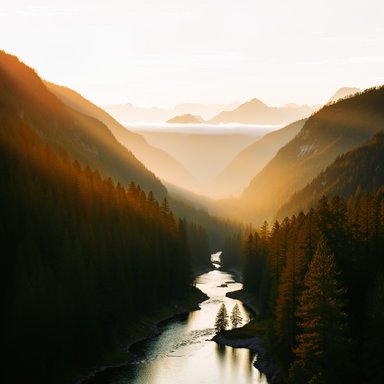} & \fcell{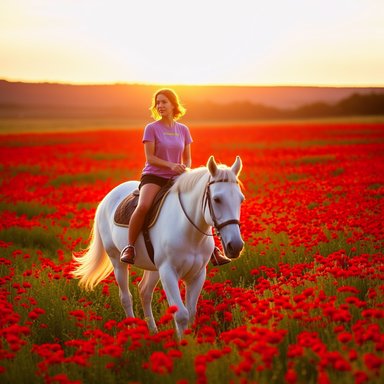} \\
\end{tabular}
\end{minipage}
\caption{Selected-setting outputs for SD3.5 Medium (left) and FLUX.2~[klein] 4B Base
(right). Rows are methods. Columns use four shared prompts: a portrait of a
woman wearing round glasses in a library; a dog to the right of a tie; a misty
mountain valley with a river at sunrise; and a woman riding a white horse
through red poppies.}
\label{fig:preview-both}
\end{figure}

\begin{figure}[!htbp]\centering
\setlength{\tabcolsep}{0pt}\renewcommand{\arraystretch}{0.6}
\begin{tabular}{@{}c@{}cccccc@{}}
 & \collab{CFG} & \collab{CFG++} & \collab{APG} & \collab{SAG} & \collab{OSEG} & \collab{PAG} \\
\rowlab{cow/stop}    & \pcell{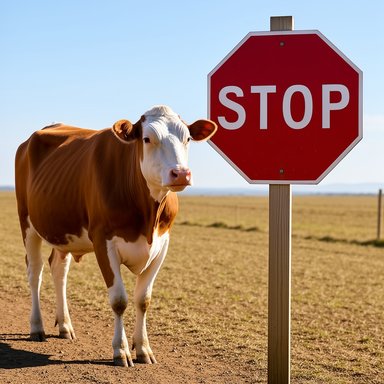}    & \pcell{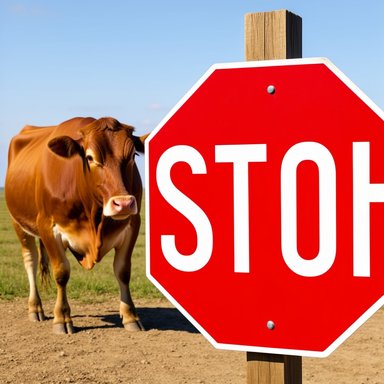}    & \pcell{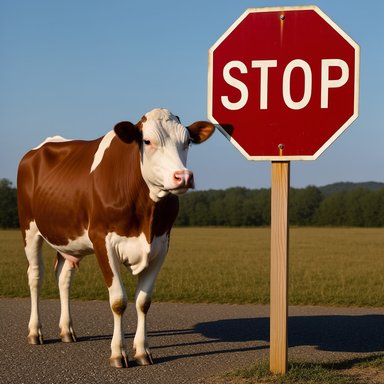}    & \pcell{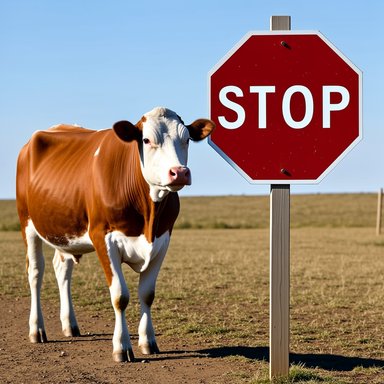}    & \pcell{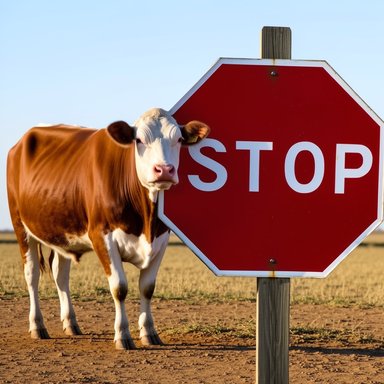}    & \pcell{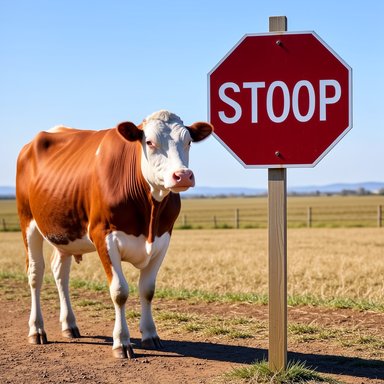} \\
\rowlab{4 frisbees}  & \pcell{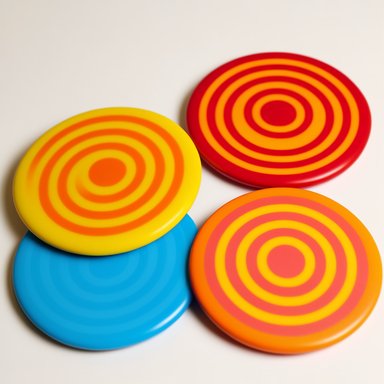}          & \pcell{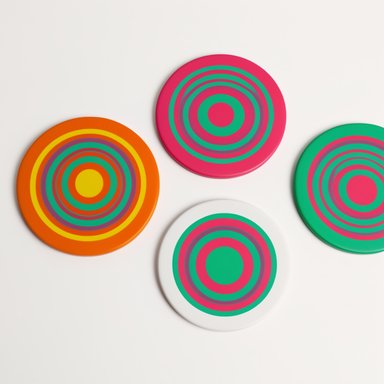}          & \pcell{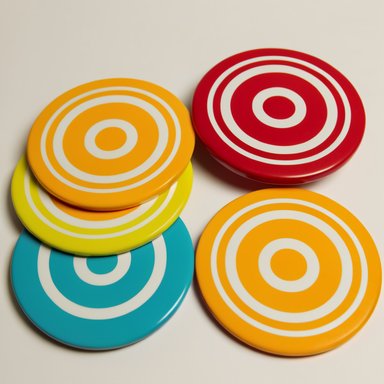}          & \pcell{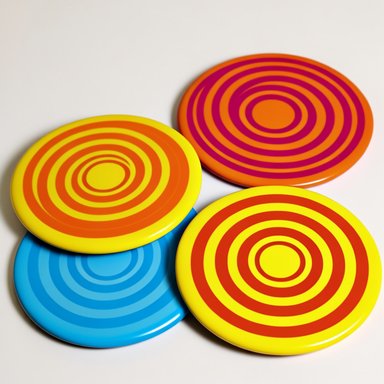}          & \pcell{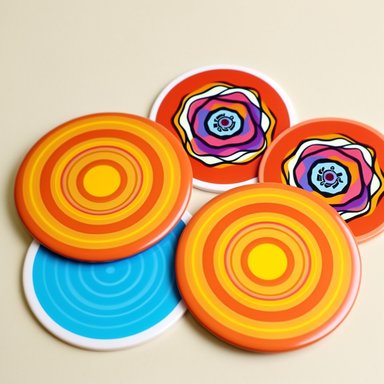}          & \pcell{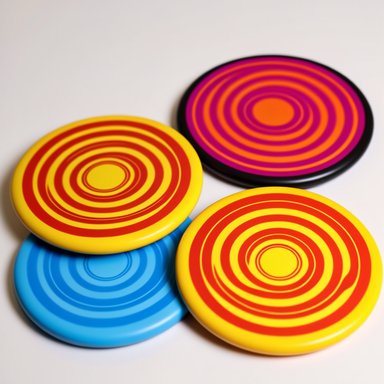} \\
\rowlab{3 giraffes}  & \pcell{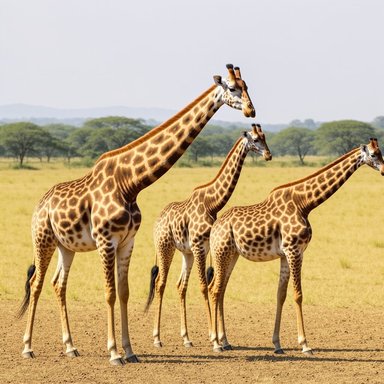}         & \pcell{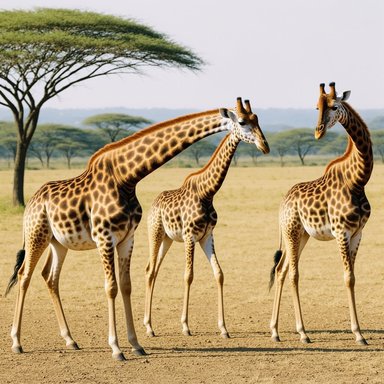}         & \pcell{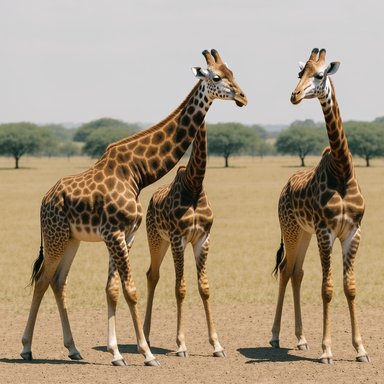}         & \pcell{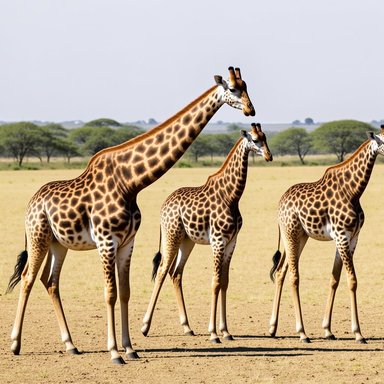}         & \pcell{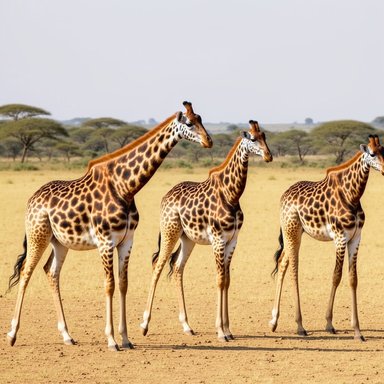}         & \pcell{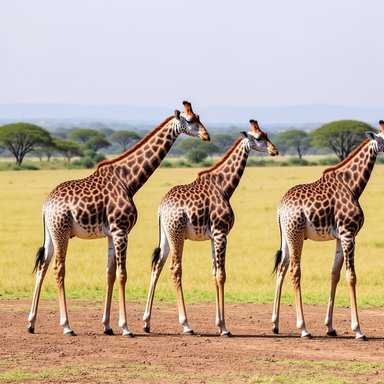} \\
\rowlab{5 workers}   & \pcell{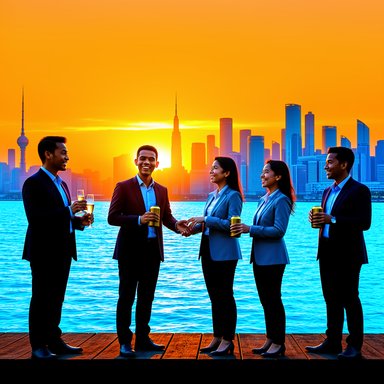}  & \pcell{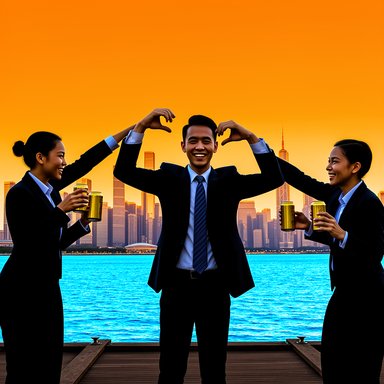}  & \pcell{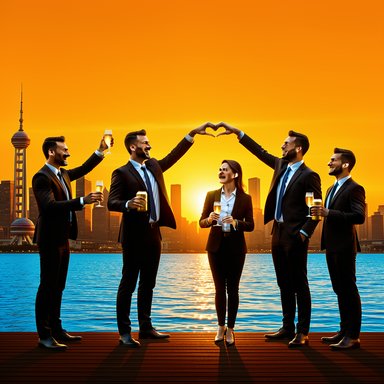}  & \pcell{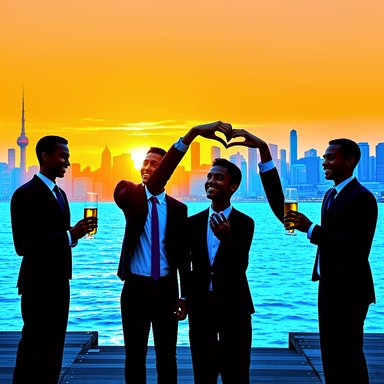}  & \pcell{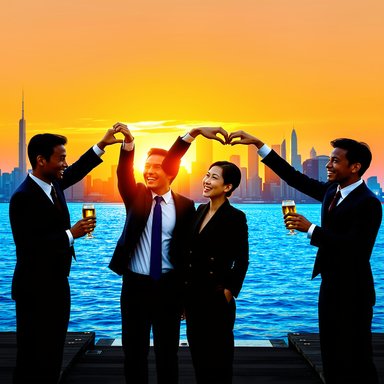}  & \pcell{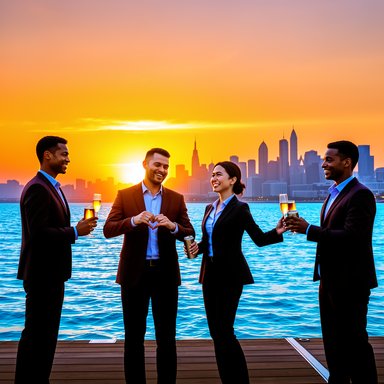} \\[3pt]
\rowlab{couch/cup}   & \pcell{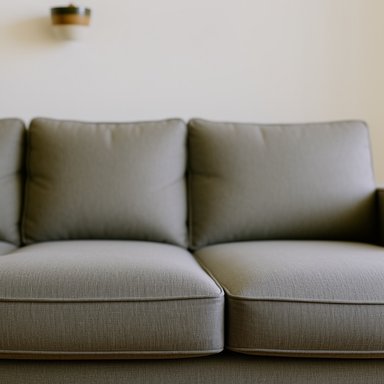}   & \pcell{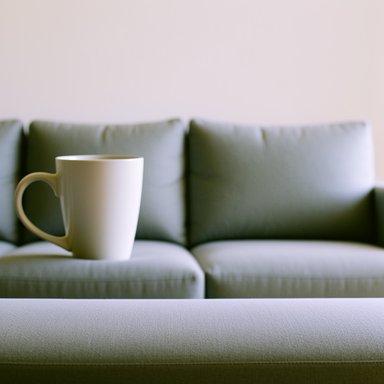}   & \pcell{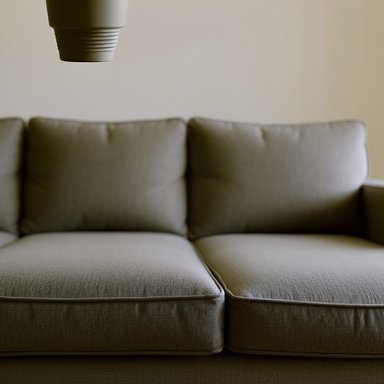}   & \pcell{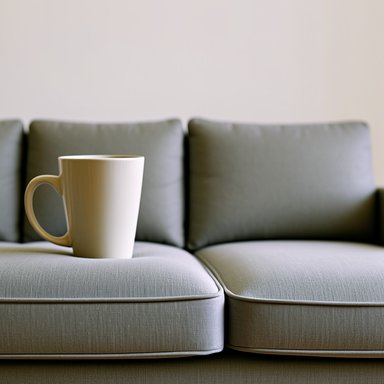}   & \pcell{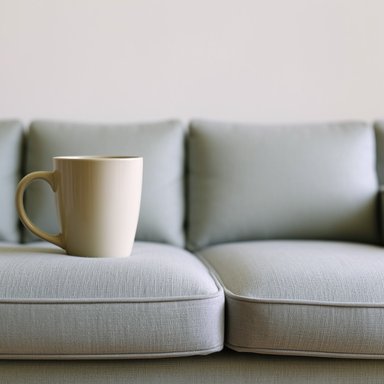}   & \pcell{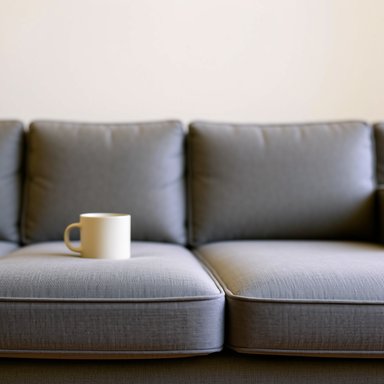} \\
\rowlab{frisbee/moto}& \pcell{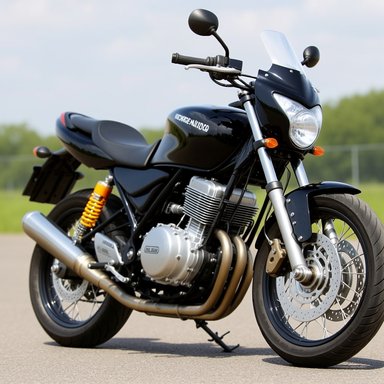}  & \pcell{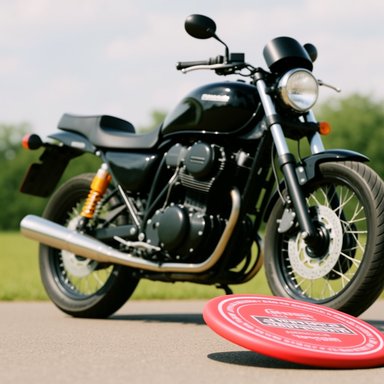}  & \pcell{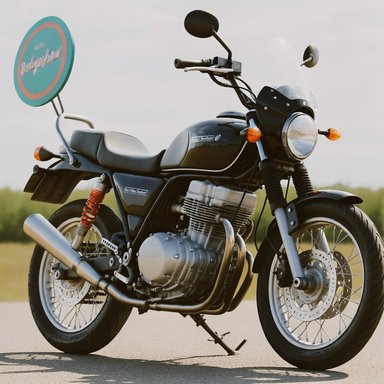}  & \pcell{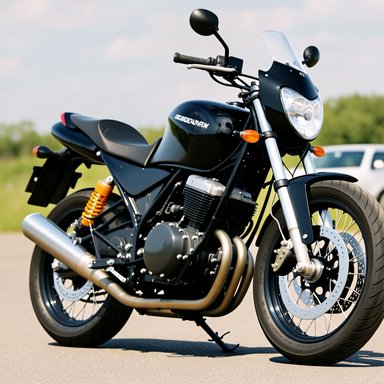}  & \pcell{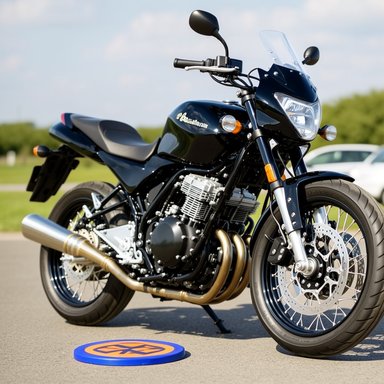}  & \pcell{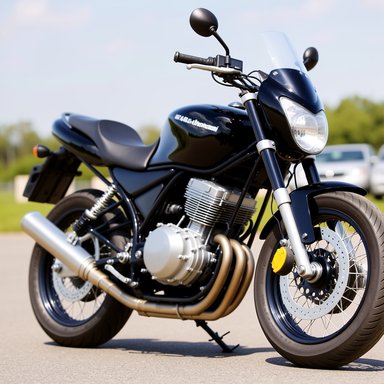} \\
\rowlab{4 vases}     & \pcell{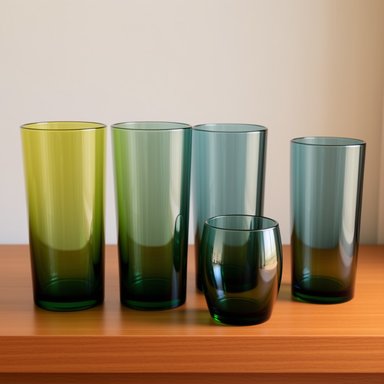}            & \pcell{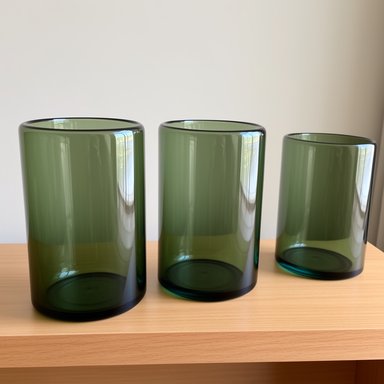}            & \pcell{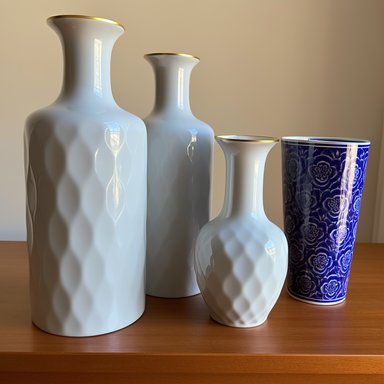}            & \pcell{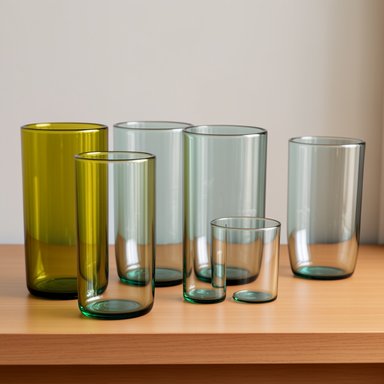}            & \pcell{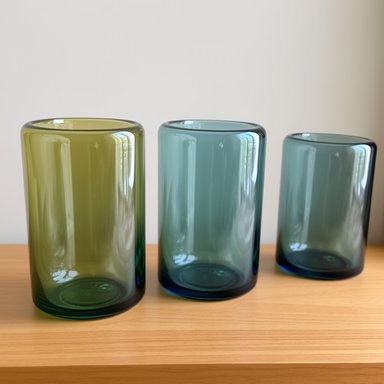}            & \pcell{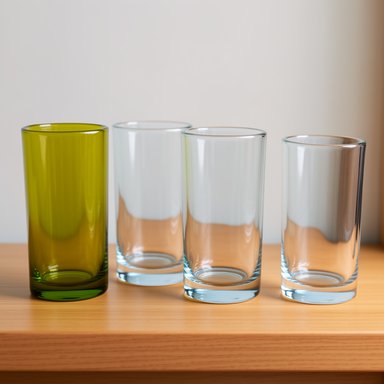} \\
\rowlab{3 bats}      & \pcell{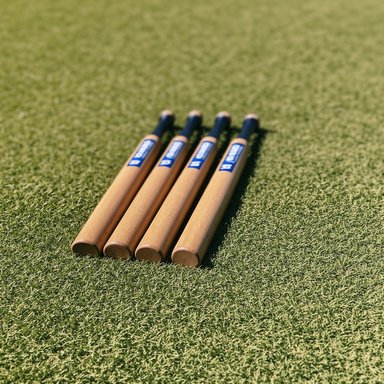}   & \pcell{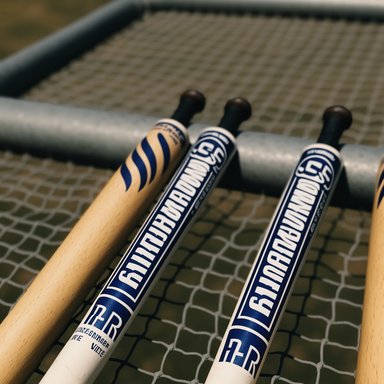}   & \pcell{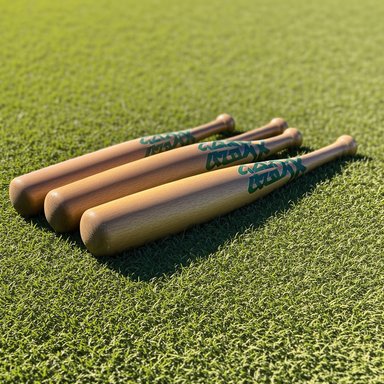}   & \pcell{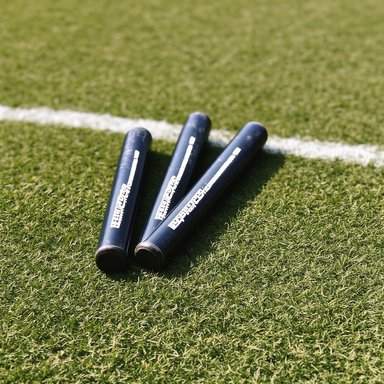}   & \pcell{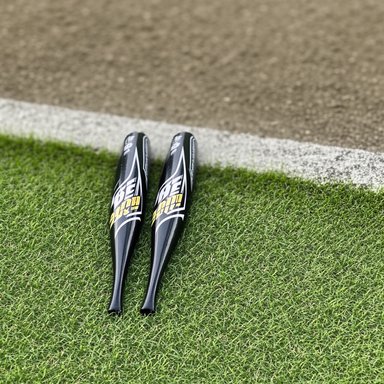}   & \pcell{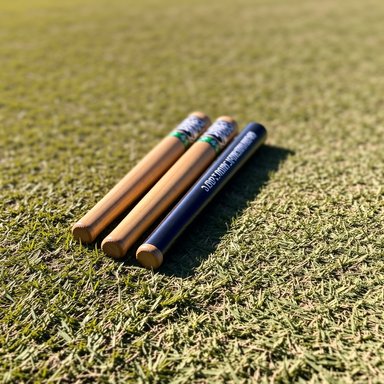} \\
\end{tabular}
\caption{Selected FLUX.2~[klein] 4B Base cases where alternatives introduce errors
relative to CFG (top) or improve its output (bottom). Columns are methods and
rows are prompts.}
\label{fig:examples-worse-better}
\end{figure}

\Cref{fig:preview-both} compares all methods at their selected settings, while
\cref{fig:examples-worse-better} shows FLUX.2~[klein] 4B Base cases in which an alternative
corrects a relation or count for one prompt but introduces an error for another.
Together, the examples illustrate prompt-level variation and do not replace the
quantitative evaluation.

\FloatBarrier

\subsection{Originally Reported Gains}
\label{sec:results-reported}

Each method was introduced with an improvement over CFG, but the evaluation
protocols are heterogeneous. \Cref{tab:reported} compares the measured property
in each original paper with our SD3.5 GenEval result. Most studies use global
quality or similarity metrics such as FID, CLIP score, Inception Score,
precision and recall, or aesthetic score. CFG-Zero* is the closest case to our
setting because it also reports T2I-CompBench on SD3/SD3.5.

The two evaluations do not produce a common ordering. Relative to the shared
SD3.5 GenEval margin, the SAG increase and the decreases of CFG++, SEG, and PAG
are resolved, while CFG-Zero*, APG, TCFG, and OSEG remain within the margin.
This comparison does not invalidate the original results: the base models,
prompts, samplers, and metrics differ. It instead shows that an improvement in
an image-quality metric does not determine the change in compositional
alignment under our protocol.

\begin{table}[!htb]\centering\small
\setlength{\tabcolsep}{4pt}
\caption{What each method's original paper measured against CFG, versus our
result. ``Comp.?'' marks whether the paper used a compositional-alignment
benchmark comparable to ours. The last column is our SD3.5 GenEval Overall
(CFG $=0.655$) with the signed difference.}
\label{tab:reported}
\begin{tabular}{@{}llcl@{}}
\toprule
Method & Original eval vs CFG & Comp.? & GenEval ($\Delta$) \\
\midrule
CFG++     & FID, CLIP, inversion (SD/SDXL)        & no  & $0.617$ \;($-0.038$) \\
CFG-Zero* & T2I-CompBench, CLIP, Aesth.\ (SD3/3.5) & yes & $0.638$ \;($-0.017$) \\
APG       & FID, recall, saturation               & no  & $0.671$ \;($+0.016$) \\
TCFG      & FID, CLIP (MS-COCO; SD3)              & no  & $0.644$ \;($-0.011$) \\
SAG       & FID, IS (ImageNet, SD1.5)             & no  & $\mathbf{0.715}$ \;($+0.060$) \\
PAG       & FID, IS (SD1.5, ImageNet)             & no  & $0.595$ \;($-0.060$) \\
SEG       & FID, CLIP (SDXL, MS-COCO)             & no  & $0.623$ \;($-0.032$) \\
OSEG      & FID / perceptual (SEG family)         & no  & $0.650$ \;($-0.005$) \\
\bottomrule
\end{tabular}
\end{table}

\FloatBarrier

\section{Discussion}
\label{sec:discussion}

Our results do not identify a universal winner among the tested guidance
methods for modern rectified-flow transformers. The methods were originally
introduced on earlier model generations, often with U-Net backbones and
different evaluation protocols. Those results describe their original
settings, but gains of the same magnitude need not transfer to current
transformer architectures.

The comparison between the two models supports this concern, although it does
not establish a general historical trend. On SD3.5, only a few improvements
from different methods exceed the shared bootstrap margins. On the newer
FLUX.2~[klein] 4B Base model, several alternatives obtain nominally higher scores, but none of
these gains exceeds the corresponding margin. At the same time, resolved
decreases relative to CFG occur for multiple methods and benchmark families.
The weaker positive effects on FLUX are consistent with the hypothesis that
guidance modifications may offer less room for improvement as base models
become stronger. Testing more model generations is necessary before this can
be treated as a broader conclusion.

This result should not be read as evidence that guidance itself has become
unimportant. CFG improves every headline benchmark substantially over no-CFG
for both models. Guidance therefore remains an important part of the tested
sampling pipelines. However, the tested alternatives, whether they modify the guidance update or construct a different weak prediction, provide no consistent gain over vanilla CFG and sometimes reduce alignment. Plain CFG therefore remains a competitive reference point.

The qualitative comparisons lead to the same mixed interpretation. Outputs
from different guided methods are often similar, but they are not identical.
For some prompts, an alternative repairs a structural defect, recovers a
missing object, or improves counting and spatial placement. For other prompts,
the same type of intervention removes an object, changes the requested count,
or weakens a relation that CFG represents correctly. These examples show how a
method can help individual generations without producing a consistent gain in
the aggregate evaluation.

\section{Limitations}
\label{sec:limitations}

Our study has several limitations. It covers two rectified-flow transformers
and a limited set of automatic alignment benchmarks. We evaluate one selected
configuration for each method and model, chosen through qualitative sweeps on a
fixed prompt set before benchmark evaluation. A wider parameter and layer
search could change individual rankings, while human evaluation and perceptual
quality metrics could reveal effects that the current benchmarks do not
measure. The shared bootstrap margins also describe uncertainty under our
prompt samples and evaluation procedure; they are not guarantees for other
prompt distributions.

Within this scope, the main distinction is between guidance and the particular
form of guidance. Taken together, these results separate the benefit of guidance from the benefit of a particular guidance rule. The former is clear in both pipelines; evidence for the latter is limited to isolated model–metric pairs and is accompanied by recurring losses.

\section*{Acknowledgments}
This research was supported in part through computational resources of HPC facilities at HSE University

\bibliographystyle{unsrtnat}
\bibliography{references}

\begin{thebibliography}{23}
\providecommand{\natexlab}[1]{#1}
\providecommand{\url}[1]{\texttt{#1}}
\expandafter\ifx\csname urlstyle\endcsname\relax
  \providecommand{\doi}[1]{doi: #1}\else
  \providecommand{\doi}{doi: \begingroup \urlstyle{rm}\Url}\fi

\bibitem[Ho and Salimans(2022)]{ho2022cfg}
Jonathan Ho and Tim Salimans.
\newblock Classifier-free diffusion guidance.
\newblock \emph{arXiv preprint arXiv:2207.12598}, 2022.

\bibitem[Esser et~al.(2024)Esser, Kulal, Blattmann, Entezari, M{\"u}ller, Saini, Levi, Lorenz, Sauer, Boesel, et~al.]{esser2024sd3}
Patrick Esser, Sumith Kulal, Andreas Blattmann, Rahim Entezari, Jonas M{\"u}ller, Harry Saini, Yam Levi, Dominik Lorenz, Axel Sauer, Frederic Boesel, et~al.
\newblock Scaling rectified flow transformers for high-resolution image synthesis.
\newblock In \emph{Forty-first international conference on machine learning}, 2024.

\bibitem[{Stability AI}(2024)]{sd35medium}
{Stability AI}.
\newblock Stable diffusion 3.5 medium.
\newblock \url{https://huggingface.co/stabilityai/stable-diffusion-3.5-medium}, 2024.

\bibitem[{Black Forest Labs}(2026)]{flux2klein}
{Black Forest Labs}.
\newblock {FLUX.2 [klein]}: Towards interactive visual intelligence.
\newblock \url{https://bfl.ai/blog/flux2-klein-towards-interactive-visual-intelligence}, 2026.

\bibitem[Ghosh et~al.(2023)Ghosh, Hajishirzi, and Schmidt]{ghosh2023geneval}
Dhruba Ghosh, Hannaneh Hajishirzi, and Ludwig Schmidt.
\newblock Geneval: An object-focused framework for evaluating text-to-image alignment.
\newblock \emph{Advances in Neural Information Processing Systems}, 36:\penalty0 52132--52152, 2023.

\bibitem[Hu et~al.(2024)Hu, Wang, Fang, Fu, Cheng, and Yu]{hu2024ella}
Xiwei Hu, Rui Wang, Yixiao Fang, Bin Fu, Pei Cheng, and Gang Yu.
\newblock Ella: Equip diffusion models with llm for enhanced semantic alignment.
\newblock \emph{arXiv preprint arXiv:2403.05135}, 2024.

\bibitem[Chang et~al.(2025)Chang, Fang, Xing, Wu, Cheng, Wang, Zeng, Yu, and Chen]{chang2026oneig}
Jingjing Chang, Yixiao Fang, Peng Xing, Shuhan Wu, Wei Cheng, Rui Wang, Xianfang Zeng, Gang Yu, and Hai-Bao Chen.
\newblock Oneig-bench: Omni-dimensional nuanced evaluation for image generation.
\newblock \emph{Advances in Neural Information Processing Systems}, 38, 2025.

\bibitem[Dhariwal and Nichol(2021)]{dhariwal2021adm}
Prafulla Dhariwal and Alexander Nichol.
\newblock Diffusion models beat gans on image synthesis.
\newblock \emph{Advances in neural information processing systems}, 34:\penalty0 8780--8794, 2021.

\bibitem[Chung et~al.(2025)Chung, Kim, Park, Nam, and Ye]{chung2024cfgpp}
Hyungjin Chung, Jeongsol Kim, Geon~Yeong Park, Hyelin Nam, and Jong~Chul Ye.
\newblock Cfg++: Manifold-constrained classifier free guidance for diffusion models.
\newblock In \emph{International Conference on Learning Representations}, 2025.

\bibitem[Fan et~al.(2025)Fan, Zheng, Yeh, and Liu]{fan2025cfgzero}
Weichen Fan, Amber~Yijia Zheng, Raymond~A Yeh, and Ziwei Liu.
\newblock Cfg-zero*: Improved classifier-free guidance for flow matching models.
\newblock \emph{arXiv preprint arXiv:2503.18886}, 2025.

\bibitem[Sadat et~al.(2025)Sadat, Hilliges, and Weber]{sadat2025apg}
Seyedmorteza Sadat, Otmar Hilliges, and Romann~M Weber.
\newblock Eliminating oversaturation and artifacts of high guidance scales in diffusion models.
\newblock In \emph{The Thirteenth International Conference on Learning Representations}, 2025.

\bibitem[Kwon et~al.(2025)Kwon, Kim, Jeong, Hsiao, and Uh]{kwon2025tcfg}
Mingi Kwon, Shinseong Kim, Jaeseok Jeong, Yi~Ting Hsiao, and Youngjung Uh.
\newblock Tcfg: Tangential damping classifier-free guidance.
\newblock In \emph{Proceedings of the Computer Vision and Pattern Recognition Conference}, pages 2620--2629, 2025.

\bibitem[Hong et~al.(2023)Hong, Lee, Jang, and Kim]{hong2023sag}
Susung Hong, Gyuseong Lee, Wooseok Jang, and Seungryong Kim.
\newblock Improving sample quality of diffusion models using self-attention guidance.
\newblock In \emph{Proceedings of the IEEE/CVF International Conference on Computer Vision}, pages 7462--7471, 2023.

\bibitem[Ahn et~al.(2024)Ahn, Cho, Min, Jang, Kim, Kim, Park, Jin, and Kim]{ahn2024pag}
Donghoon Ahn, Hyoungwon Cho, Jaewon Min, Wooseok Jang, Jungwoo Kim, SeonHwa Kim, Hyun~Hee Park, Kyong~Hwan Jin, and Seungryong Kim.
\newblock Self-rectifying diffusion sampling with perturbed-attention guidance.
\newblock In \emph{European Conference on Computer Vision}, pages 1--17. Springer, 2024.

\bibitem[Hong(2024)]{hong2024seg}
Susung Hong.
\newblock Smoothed energy guidance: Guiding diffusion models with reduced energy curvature of attention.
\newblock \emph{Advances in Neural Information Processing Systems}, 37:\penalty0 66743--66772, 2024.

\bibitem[Fahim et~al.(2026)Fahim, Saqib, and Gil]{fahim2026oseg}
Masud An Nur~Islam Fahim, Nazmus Saqib, and Joon-Min Gil.
\newblock {OSEG}: Improving diffusion sampling through orthogonal smoothed energy guidance.
\newblock In \emph{Proceedings of the IEEE/CVF Winter Conference on Applications of Computer Vision}, pages 5996--6005, 2026.

\bibitem[Kynk{\"a}{\"a}nniemi et~al.(2024)Kynk{\"a}{\"a}nniemi, Aittala, Karras, Laine, Aila, and Lehtinen]{kynkaanniemi2024interval}
Tuomas Kynk{\"a}{\"a}nniemi, Miika Aittala, Tero Karras, Samuli Laine, Timo Aila, and Jaakko Lehtinen.
\newblock Applying guidance in a limited interval improves sample and distribution quality in diffusion models.
\newblock \emph{Advances in Neural Information Processing Systems}, 37:\penalty0 122458--122483, 2024.

\bibitem[Sadat et~al.(2024)Sadat, Buhmann, Bradley, Hilliges, and Weber]{sadat2024cads}
Seyedmorteza Sadat, Jakob Buhmann, Derek Bradley, Otmar Hilliges, and Romann Weber.
\newblock Cads: Unleashing the diversity of diffusion models through condition-annealed sampling.
\newblock In \emph{International Conference on Learning Representations}, 2024.

\bibitem[Karras et~al.(2024)Karras, Aittala, Kynk{\"a}{\"a}nniemi, Lehtinen, Aila, and Laine]{karras2024autoguidance}
Tero Karras, Miika Aittala, Tuomas Kynk{\"a}{\"a}nniemi, Jaakko Lehtinen, Timo Aila, and Samuli Laine.
\newblock Guiding a diffusion model with a bad version of itself.
\newblock \emph{Advances in Neural Information Processing Systems}, 37:\penalty0 52996--53021, 2024.

\bibitem[Hu et~al.(2023)Hu, Liu, Kasai, Wang, Ostendorf, Krishna, and Smith]{hu2023tifa}
Yushi Hu, Benlin Liu, Jungo Kasai, Yizhong Wang, Mari Ostendorf, Ranjay Krishna, and Noah~A Smith.
\newblock Tifa: Accurate and interpretable text-to-image faithfulness evaluation with question answering.
\newblock In \emph{Proceedings of the IEEE/CVF International Conference on Computer Vision}, pages 20406--20417, 2023.

\bibitem[Huang et~al.(2023)Huang, Sun, Xie, Li, and Liu]{huang2023t2icompbench}
Kaiyi Huang, Kaiyue Sun, Enze Xie, Zhenguo Li, and Xihui Liu.
\newblock T2i-compbench: A comprehensive benchmark for open-world compositional text-to-image generation.
\newblock \emph{Advances in Neural Information Processing Systems}, 36:\penalty0 78723--78747, 2023.

\bibitem[Xu et~al.(2023)Xu, Liu, Wu, Tong, Li, Ding, Tang, and Dong]{xu2023imagereward}
Jiazheng Xu, Xiao Liu, Yuchen Wu, Yuxuan Tong, Qinkai Li, Ming Ding, Jie Tang, and Yuxiao Dong.
\newblock Imagereward: Learning and evaluating human preferences for text-to-image generation.
\newblock \emph{Advances in Neural Information Processing Systems}, 36:\penalty0 15903--15935, 2023.

\bibitem[Ahn et~al.(2025)Ahn, Kang, Lee, Kim, Min, Jang, Lee, Paul, Hong, and Kim]{ahn2025headhunter}
Donghoon Ahn, Jiwon Kang, Sanghyun Lee, Minjae Kim, Jaewon Min, Wooseok Jang, Sangwu Lee, Sayak Paul, Susung Hong, and Seungryong Kim.
\newblock Where and how to perturb: On the design of perturbation guidance in diffusion and flow models.
\newblock \emph{Advances in Neural Information Processing Systems}, 38:\penalty0 128348--128396, 2025.

\end{thebibliography}

\appendix

\section{Hyperparameter Settings}
\label{sec:app-hparams}
Final per-method configurations selected by the qualitative procedure described
in the main paper, per base model. Both runs use a single forward pass per
prediction; attention-perturbation methods add one extra pass. Knobs not listed
take their pipeline defaults.

\begin{table}[!htb]\centering\small
\caption{Final hyperparameters per method, SD3.5 Medium (25 steps).}
\label{tab:hparams-sd35}
\begin{tabular}{ll}
\toprule
Method & Settings \\
\midrule
no-CFG    & $w=1$ \\
CFG       & $w=4.0$ \\
CFG++     & $\lambda=0.75$ \\
CFG-Zero* & $w=3.5$, zero-init steps $=0$ \\
APG       & $w=7$, step radius $=20$, momentum $=-0.4$ \\
TCFG      & $w=6$, rank $=1$ \\
SAG       & $w=5$, sag scale $=0.4$, layer d0 \\
SEG       & $w=3$, seg scale $=3$, blur $\sigma=10$, layer d8 \\
OSEG      & $w=4$, oseg scale $=2.5$, blur $\sigma=10$, layer d8 \\
PAG       & $w=2.5$, pag scale $=1.0$, layer s0 \\
\bottomrule
\end{tabular}
\end{table}

\begin{table}[!htb]\centering\small
\caption{Final hyperparameters per method, FLUX.2~[klein] 4B Base (30 steps).}
\label{tab:hparams-flux}
\begin{tabular}{ll}
\toprule
Method & Settings \\
\midrule
no-CFG    & guidance off \\
CFG       & $w=7.0$ \\
CFG++     & $w=1.2$ \\
CFG-Zero* & $w=7.0$, zero-init steps $=2$ \\
APG       & $w=10$, momentum $=-0.5$, $\eta=0$, step radius $=0$ \\
TCFG      & $w=12$, rank $=1$ \\
SAG       & $w=5$, sag scale $=0.4$, blur $\sigma=3$, layer d0 \\
SEG       & $w=4$, seg scale $=2.0$, blur $\sigma=10$, layers d4,\,s2 \\
OSEG      & $w=4$, oseg scale $=0.75$, blur $\sigma=5$, layer s2 \\
PAG       & $w=4.5$, pag scale $=1.5$, layers d4,\,s1 \\
\bottomrule
\end{tabular}
\end{table}

\FloatBarrier
\section{Detailed Benchmark Results}
\label{sec:app-details}
The tables below provide the GenEval and DPG-Bench category breakdowns for both
models. All evaluated OneIG families already appear in the headline table of the
main paper and are not repeated here. The colors provide a descriptive comparison with
CFG: gray indicates near-parity, green/red indicate the direction of change, and
stronger shades indicate larger differences. They are visual guides and do not
denote statistical significance for individual categories.

\begin{table}[!htbp]\centering\tiny
\setlength{\tabcolsep}{1.1pt}
\renewcommand{\arraystretch}{1.95}
\caption{GenEval category scores for both models. Overall is the macro average.}
\label{tab:geneval-both}
\resizebox{\textwidth}{!}{%
\begin{tabular}{@{}l*{7}{c}@{\hspace{2pt}{\color{black!30}\vrule width 0.35pt}\hspace{2pt}}*{7}{c}@{}}
\toprule
& \multicolumn{7}{c}{\textbf{SD3.5 Medium}}
& \multicolumn{7}{c}{\textbf{FLUX.2~[klein] 4B Base}} \\
\cmidrule(lr){2-8}\cmidrule(lr){9-15}
Method & single & two & count & color & pos & attr & \textbf{Overall}
& single & two & count & color & pos & attr & \textbf{Overall} \\
\midrule
no-CFG & \resultlossstrong{0.767} & \resultlossstrong{0.500} & \resultlossstrong{0.194} & \resultlossstrong{0.419} & \resultlossstrong{0.097} & \resultlossstrong{0.226} & \resultlossstrong{0.367} & \resultlossstrong{0.800} & \resultlossstrong{0.242} & \resultlossstrong{0.225} & \resultlossstrong{0.596} & \resultlossstrong{0.120} & \resultlossstrong{0.090} & \resultlossstrong{0.346} \\
CFG & \resulttie{\textbf{1.000}} & \resulttie{0.833} & \resulttie{0.484} & \resulttie{0.806} & \resulttie{0.226} & \resulttie{0.581} & \resulttie{0.655} & \resulttie{0.988} & \resulttie{\textbf{0.889}} & \resulttie{\textbf{0.788}} & \resulttie{0.862} & \resulttie{0.560} & \resulttie{\textbf{0.630}} & \resulttie{0.786} \\
CFG++ & \resulttie{\textbf{1.000}} & \resulttie{0.833} & \resulttie{0.484} & \resultlossstrong{0.742} & \resultlossstrong{0.161} & \resultlossstrong{0.484} & \resultloss{0.617} & \resulttie{0.988} & \resultloss{0.848} & \resultlossstrong{0.550} & \resulttie{0.840} & \resultlossstrong{0.450} & \resultlossstrong{0.550} & \resultlossstrong{0.704} \\
CFG-Zero* & \resulttie{\textbf{1.000}} & \resultlossstrong{0.733} & \resulttie{0.484} & \resultgain{\textbf{0.839}} & \resultgain{0.258} & \resultlossstrong{0.516} & \resulttie{0.638} & \resulttie{\textbf{1.000}} & \resultlossstrong{0.818} & \resultlossstrong{0.663} & \resulttie{\textbf{0.883}} & \resultlossstrong{0.410} & \resultloss{0.580} & \resultlossstrong{0.726} \\
APG & \resulttie{\textbf{1.000}} & \resulttie{0.833} & \resultgainstrong{0.581} & \resultloss{0.774} & \resultgain{0.258} & \resulttie{0.581} & \resulttie{0.671} & \resulttie{\textbf{1.000}} & \resulttie{\textbf{0.889}} & \resultloss{0.738} & \resulttie{\textbf{0.883}} & \resultgainstrong{\textbf{0.620}} & \resulttie{0.620} & \resulttie{\textbf{0.792}} \\
TCFG & \resulttie{\textbf{1.000}} & \resultlossstrong{0.767} & \resultgainstrong{0.548} & \resultlossstrong{0.677} & \resultgainstrong{\textbf{0.290}} & \resulttie{0.581} & \resulttie{0.644} & \resulttie{0.988} & \resultloss{0.859} & \resulttie{\textbf{0.788}} & \resulttie{0.851} & \resulttie{0.550} & \resulttie{0.610} & \resulttie{0.774} \\
SAG & \resulttie{\textbf{1.000}} & \resultgainstrong{\textbf{0.967}} & \resultgainstrong{\textbf{0.645}} & \resultloss{0.774} & \resulttie{0.226} & \resultgainstrong{\textbf{0.677}} & \resultgainstrong{\textbf{0.715}} & \resulttie{0.988} & \resulttie{0.869} & \resultlossstrong{0.725} & \resulttie{0.862} & \resulttie{0.550} & \resultloss{0.580} & \resulttie{0.762} \\
SEG & \resultloss{0.967} & \resulttie{0.833} & \resultgainstrong{0.548} & \resultlossstrong{0.710} & \resultlossstrong{0.161} & \resultlossstrong{0.516} & \resultloss{0.623} & \resulttie{\textbf{1.000}} & \resultlossstrong{0.808} & \resultlossstrong{0.663} & \resulttie{0.872} & \resultlossstrong{0.490} & \resultlossstrong{0.510} & \resultlossstrong{0.724} \\
OSEG & \resultloss{0.967} & \resultgain{0.867} & \resultgainstrong{0.613} & \resultlossstrong{0.677} & \resultloss{0.194} & \resulttie{0.581} & \resulttie{0.650} & \resulttie{\textbf{1.000}} & \resulttie{0.879} & \resultlossstrong{0.700} & \resulttie{0.851} & \resultloss{0.510} & \resultlossstrong{0.520} & \resultloss{0.743} \\
PAG & \resulttie{\textbf{1.000}} & \resultlossstrong{0.700} & \resultlossstrong{0.419} & \resultlossstrong{0.742} & \resulttie{0.226} & \resultlossstrong{0.484} & \resultlossstrong{0.595} & \resulttie{0.988} & \resultloss{0.859} & \resultloss{0.738} & \resultlossstrong{0.787} & \resultloss{0.520} & \resultlossstrong{0.520} & \resultloss{0.735} \\
\bottomrule
\end{tabular}%
}
\end{table}

\begin{table}[!htbp]\centering\tiny
\setlength{\tabcolsep}{1.3pt}
\renewcommand{\arraystretch}{1.50}
\caption{DPG-Bench L1 scores for both models. DPG is mean$\times$100.}
\label{tab:dpg-l1-both}
\resizebox{\textwidth}{!}{%
\begin{tabular}{@{}l*{6}{c}@{\hspace{2pt}{\color{black!30}\vrule width 0.35pt}\hspace{2pt}}*{6}{c}@{}}
\toprule
& \multicolumn{6}{c}{\textbf{SD3.5 Medium}}
& \multicolumn{6}{c}{\textbf{FLUX.2~[klein] 4B Base}} \\
\cmidrule(lr){2-7}\cmidrule(lr){8-13}
Method & DPG & Attr & Entity & Other & Global & Relation
& DPG & Attr & Entity & Other & Global & Relation \\
\midrule
no-CFG & \resultlossstrong{73.51} & \resultlossstrong{79.87} & \resultlossstrong{82.48} & \resultlossstrong{68.89} & \resultlossstrong{77.57} & \resultlossstrong{91.65} & \resultlossstrong{73.22} & \resultlossstrong{83.70} & \resultlossstrong{82.13} & \resultlossstrong{61.60} & \resulttie{81.76} & \resultloss{91.80} \\
CFG & \resulttie{84.35} & \resulttie{88.47} & \resulttie{90.58} & \resulttie{\textbf{84.44}} & \resulttie{83.18} & \resulttie{94.03} & \resulttie{83.30} & \resulttie{89.82} & \resulttie{90.15} & \resulttie{81.60} & \resulttie{81.46} & \resulttie{93.19} \\
CFG++ & \resulttie{84.29} & \resultloss{87.07} & \resulttie{90.08} & \resultlossstrong{81.11} & \resultgainstrong{\textbf{85.05}} & \resulttie{94.03} & \resulttie{83.35} & \resulttie{89.62} & \resulttie{90.03} & \resulttie{81.20} & \resulttie{82.07} & \resulttie{93.73} \\
CFG-Zero* & \resulttie{84.19} & \resulttie{88.47} & \resulttie{90.58} & \resultlossstrong{80.00} & \resultlossstrong{79.44} & \resulttie{93.68} & \resulttie{82.91} & \resulttie{89.64} & \resulttie{90.03} & \resultgain{\textbf{83.20}} & \resulttie{81.16} & \resulttie{93.54} \\
APG & \resulttie{84.54} & \resulttie{\textbf{88.82}} & \resulttie{\textbf{90.99}} & \resultlossstrong{80.00} & \resulttie{83.18} & \resulttie{\textbf{94.75}} & \resulttie{\textbf{83.50}} & \resulttie{89.48} & \resulttie{\textbf{90.40}} & \resultlossstrong{78.00} & \resulttie{81.46} & \resulttie{93.93} \\
TCFG & \resultloss{83.45} & \resulttie{87.89} & \resultloss{89.53} & \resulttie{\textbf{84.44}} & \resulttie{83.18} & \resulttie{94.27} & \resulttie{83.00} & \resulttie{89.74} & \resulttie{90.10} & \resulttie{82.00} & \resulttie{81.16} & \resulttie{93.27} \\
SAG & \resulttie{84.41} & \resulttie{88.77} & \resulttie{90.53} & \resultlossstrong{78.89} & \resulttie{83.18} & \resulttie{94.27} & \resulttie{83.33} & \resulttie{\textbf{89.90}} & \resulttie{90.10} & \resultloss{80.00} & \resulttie{80.85} & \resulttie{93.46} \\
SEG & \resulttie{84.03} & \resulttie{87.83} & \resulttie{89.78} & \resultloss{83.33} & \resultlossstrong{80.37} & \resulttie{93.68} & \resulttie{83.21} & \resulttie{89.20} & \resulttie{90.36} & \resultloss{80.00} & \resultgainstrong{\textbf{83.89}} & \resultgain{\textbf{94.08}} \\
OSEG & \resulttie{\textbf{84.61}} & \resulttie{87.65} & \resulttie{90.68} & \resultlossstrong{80.00} & \resultlossstrong{81.31} & \resulttie{\textbf{94.75}} & \resulttie{83.11} & \resulttie{89.52} & \resulttie{89.77} & \resulttie{82.00} & \resultgain{82.67} & \resulttie{93.62} \\
PAG & \resultlossstrong{82.56} & \resultloss{87.24} & \resultloss{88.92} & \resultlossstrong{77.78} & \resultloss{82.24} & \resulttie{94.27} & \resulttie{83.35} & \resulttie{89.66} & \resulttie{90.03} & \resultlossstrong{79.20} & \resulttie{81.76} & \resulttie{93.35} \\
\bottomrule
\end{tabular}%
}
\end{table}

\FloatBarrier

\begin{table}[!ht]\centering\tiny
\setlength{\tabcolsep}{1.5pt}
\renewcommand{\arraystretch}{1.25}
\caption{DPG-Bench L2 attribute scores for both models.}
\label{tab:dpg-l2-attr-both}
\resizebox{\textwidth}{!}{%
\begin{tabular}{@{}l*{5}{c}@{\hspace{2pt}{\color{black!30}\vrule width 0.35pt}\hspace{2pt}}*{5}{c}@{}}
\toprule
& \multicolumn{5}{c}{\textbf{SD3.5 Medium}}
& \multicolumn{5}{c}{\textbf{FLUX.2~[klein] 4B Base}} \\
\cmidrule(lr){2-6}\cmidrule(lr){7-11}
Method & color & other & shape & size & texture
& color & other & shape & size & texture \\
\midrule
no-CFG & \resultlossstrong{83.46} & \resultlossstrong{78.78} & \resultloss{78.46} & \resultlossstrong{66.67} & \resultlossstrong{78.60} & \resultlossstrong{87.05} & \resultlossstrong{82.71} & \resultlossstrong{75.54} & \resultlossstrong{69.01} & \resultlossstrong{83.49} \\
CFG & \resulttie{92.93} & \resulttie{87.50} & \resulttie{80.00} & \resulttie{70.97} & \resulttie{\textbf{87.64}} & \resulttie{93.18} & \resulttie{\textbf{87.47}} & \resulttie{82.97} & \resulttie{76.86} & \resulttie{89.90} \\
CFG++ & \resultloss{91.73} & \resultlossstrong{85.47} & \resultloss{78.46} & \resultgainstrong{\textbf{77.42}} & \resultlossstrong{85.06} & \resulttie{\textbf{93.68}} & \resulttie{86.92} & \resulttie{83.41} & \resultgain{78.10} & \resultloss{88.75} \\
CFG-Zero* & \resulttie{93.53} & \resulttie{86.92} & \resultgain{81.54} & \resultgainstrong{74.19} & \resultloss{86.53} & \resulttie{93.33} & \resultloss{86.47} & \resultgain{\textbf{83.84}} & \resulttie{76.86} & \resulttie{89.59} \\
APG & \resultgain{93.83} & \resulttie{87.50} & \resultgainstrong{\textbf{84.62}} & \resultgain{72.04} & \resulttie{86.90} & \resulttie{92.98} & \resulttie{87.25} & \resultgain{\textbf{83.84}} & \resultloss{75.62} & \resulttie{89.29} \\
TCFG & \resultloss{92.03} & \resultgain{88.37} & \resulttie{80.00} & \resulttie{70.97} & \resultloss{86.35} & \resulttie{93.23} & \resulttie{87.36} & \resulttie{82.53} & \resulttie{77.27} & \resulttie{89.66} \\
SAG & \resultgain{\textbf{94.29}} & \resultgain{\textbf{88.95}} & \resultgainstrong{\textbf{84.62}} & \resulttie{70.97} & \resultlossstrong{85.42} & \resulttie{93.43} & \resulttie{86.70} & \resultlossstrong{80.35} & \resultgainstrong{\textbf{80.17}} & \resulttie{\textbf{90.14}} \\
SEG & \resulttie{92.33} & \resultlossstrong{85.47} & \resultgainstrong{83.08} & \resulttie{70.97} & \resulttie{87.27} & \resulttie{92.88} & \resultloss{86.47} & \resulttie{82.97} & \resulttie{76.45} & \resultloss{88.99} \\
OSEG & \resulttie{92.63} & \resulttie{87.50} & \resulttie{80.00} & \resultgainstrong{73.12} & \resultlossstrong{85.06} & \resulttie{93.03} & \resulttie{87.36} & \resulttie{82.97} & \resulttie{77.69} & \resulttie{89.11} \\
PAG & \resultloss{91.88} & \resultloss{86.05} & \resultgainstrong{83.08} & \resultloss{69.89} & \resultlossstrong{85.79} & \resulttie{92.77} & \resulttie{87.14} & \resulttie{83.41} & \resultgainstrong{\textbf{80.17}} & \resulttie{89.53} \\
\bottomrule
\end{tabular}%
}
\end{table}

\begin{table}[!ht]\centering\tiny
\setlength{\tabcolsep}{1.5pt}
\renewcommand{\arraystretch}{1.25}
\caption{DPG-Bench L2 non-attribute scores for SD3.5 Medium.}
\label{tab:sd35-dpg-l2-rest}
\resizebox{0.92\textwidth}{!}{%
\begin{tabular}{@{}lrrrrrrrr@{}}
\toprule
Method & part & state & whole & global & count & text & non-sp. & spatial \\
\midrule
no-CFG & \resultlossstrong{81.05} & \resultlossstrong{77.54} & \resultlossstrong{83.49} & \resultlossstrong{77.57} & \resultlossstrong{67.61} & \resultlossstrong{73.68} & \resultlossstrong{85.71} & \resultlossstrong{92.07} \\
CFG & \resulttie{86.27} & \resulttie{85.14} & \resulttie{91.97} & \resulttie{83.18} & \resulttie{\textbf{83.10}} & \resulttie{\textbf{89.47}} & \resulttie{94.64} & \resulttie{93.99} \\
CFG++ & \resultlossstrong{79.74} & \resultlossstrong{82.97} & \resulttie{92.36} & \resultgainstrong{\textbf{85.05}} & \resultloss{81.69} & \resultlossstrong{78.95} & \resultgainstrong{\textbf{96.43}} & \resulttie{93.86} \\
CFG-Zero* & \resultlossstrong{84.31} & \resulttie{84.42} & \resulttie{92.29} & \resultlossstrong{79.44} & \resultlossstrong{77.46} & \resulttie{\textbf{89.47}} & \resultlossstrong{92.86} & \resulttie{93.73} \\
APG & \resultlossstrong{84.31} & \resulttie{85.51} & \resulttie{\textbf{92.61}} & \resulttie{83.18} & \resultlossstrong{77.46} & \resulttie{\textbf{89.47}} & \resultlossstrong{91.07} & \resultgain{\textbf{95.01}} \\
TCFG & \resultlossstrong{83.01} & \resultlossstrong{82.61} & \resulttie{91.39} & \resulttie{83.18} & \resulttie{\textbf{83.10}} & \resulttie{\textbf{89.47}} & \resultlossstrong{92.86} & \resulttie{94.37} \\
SAG & \resulttie{85.62} & \resultgain{\textbf{86.23}} & \resulttie{91.78} & \resulttie{83.18} & \resultlossstrong{77.46} & \resultlossstrong{84.21} & \resultlossstrong{92.86} & \resulttie{94.37} \\
SEG & \resultloss{84.97} & \resultlossstrong{81.88} & \resulttie{91.65} & \resultlossstrong{80.37} & \resultloss{81.69} & \resulttie{\textbf{89.47}} & \resultlossstrong{91.07} & \resulttie{93.86} \\
OSEG & \resulttie{\textbf{86.93}} & \resulttie{85.51} & \resulttie{91.97} & \resultlossstrong{81.31} & \resultlossstrong{78.87} & \resultlossstrong{84.21} & \resulttie{94.64} & \resulttie{94.76} \\
PAG & \resultlossstrong{81.70} & \resultlossstrong{82.97} & \resultloss{90.69} & \resultloss{82.24} & \resultlossstrong{74.65} & \resulttie{\textbf{89.47}} & \resultlossstrong{91.07} & \resulttie{94.50} \\
\bottomrule
\end{tabular}%
}
\end{table}

\FloatBarrier

\begin{table}[!ht]\centering\tiny
\setlength{\tabcolsep}{1.5pt}
\renewcommand{\arraystretch}{1.25}
\caption{DPG-Bench L2 non-attribute scores for FLUX.2~[klein] 4B Base.}
\label{tab:flux-dpg-l2-rest}
\resizebox{0.92\textwidth}{!}{%
\begin{tabular}{@{}lrrrrrrrr@{}}
\toprule
Method & part & state & whole & global & count & text & non-sp. & spatial \\
\midrule
no-CFG & \resultlossstrong{80.47} & \resultlossstrong{81.52} & \resultlossstrong{82.43} & \resulttie{81.76} & \resultlossstrong{57.50} & \resultlossstrong{78.00} & \resultlossstrong{84.91} & \resultloss{92.25} \\
CFG & \resulttie{84.77} & \resulttie{84.69} & \resulttie{91.82} & \resulttie{81.46} & \resulttie{78.50} & \resulttie{94.00} & \resulttie{86.79} & \resulttie{93.61} \\
CFG++ & \resulttie{85.16} & \resulttie{84.79} & \resulttie{91.61} & \resulttie{82.07} & \resultloss{77.50} & \resultgainstrong{\textbf{96.00}} & \resulttie{87.42} & \resulttie{94.15} \\
CFG-Zero* & \resultgain{85.74} & \resulttie{\textbf{85.11}} & \resulttie{91.48} & \resulttie{81.16} & \resultgain{\textbf{80.00}} & \resultgainstrong{\textbf{96.00}} & \resulttie{86.79} & \resulttie{93.98} \\
APG & \resultgain{\textbf{85.94}} & \resulttie{84.16} & \resulttie{\textbf{92.14}} & \resulttie{81.46} & \resultlossstrong{74.00} & \resulttie{94.00} & \resultloss{85.53} & \resultgain{\textbf{94.48}} \\
TCFG & \resulttie{83.98} & \resulttie{84.16} & \resulttie{91.95} & \resulttie{81.16} & \resulttie{79.00} & \resulttie{94.00} & \resulttie{86.16} & \resulttie{93.73} \\
SAG & \resulttie{84.38} & \resulttie{83.95} & \resulttie{91.95} & \resulttie{80.85} & \resultlossstrong{76.00} & \resultgainstrong{\textbf{96.00}} & \resultlossstrong{84.91} & \resulttie{94.02} \\
SEG & \resultloss{83.59} & \resulttie{\textbf{85.11}} & \resulttie{\textbf{92.14}} & \resultgainstrong{\textbf{83.89}} & \resultlossstrong{76.50} & \resulttie{94.00} & \resultgainstrong{\textbf{88.68}} & \resulttie{94.44} \\
OSEG & \resultloss{83.20} & \resulttie{84.16} & \resulttie{91.61} & \resultgain{82.67} & \resulttie{78.50} & \resultgainstrong{\textbf{96.00}} & \resulttie{87.42} & \resulttie{94.02} \\
PAG & \resulttie{84.57} & \resultloss{83.74} & \resulttie{91.88} & \resulttie{81.76} & \resultlossstrong{75.00} & \resultgainstrong{\textbf{96.00}} & \resultloss{85.53} & \resulttie{93.86} \\
\bottomrule
\end{tabular}%
}
\end{table}

\FloatBarrier

\section{Hyperparameter Sweep Grids}
\label{sec:app-sweeps}
These qualitative sweeps were used only to select stable per-method settings
before running the benchmarks.

\begin{figure}[!htbp]\centering
\captionsetup{font=large,labelfont=bf,justification=centering,singlelinecheck=false}
\setlength{\tabcolsep}{0pt}\renewcommand{\arraystretch}{0}
\begin{tabular}{@{}c@{}cccc@{}c@{}}
 & \cfgcollab{1.5} & \cfgcollab{5} & \cfgcollab{12} & \cfgcollab{18} & \cfgspacer \\
\cfgrowlab{CFG}  & \cfgcell{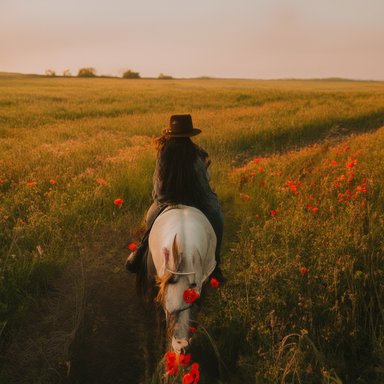}  & \cfgcell{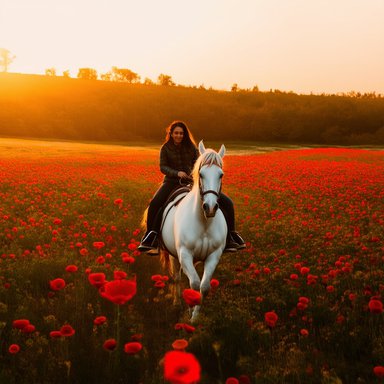}  & \cfgcell{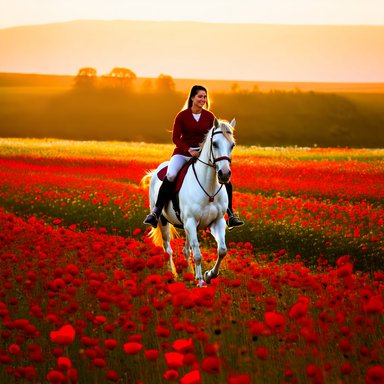}  & \cfgcell{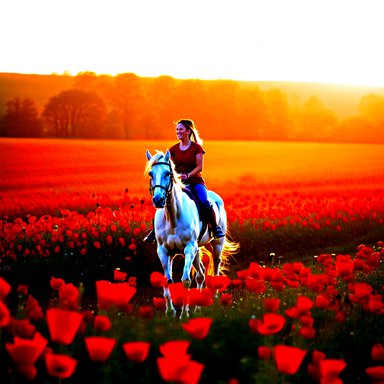} & \cfgspacer \\[-8pt]
\cfgrowlab{APG}  & \cfgcell{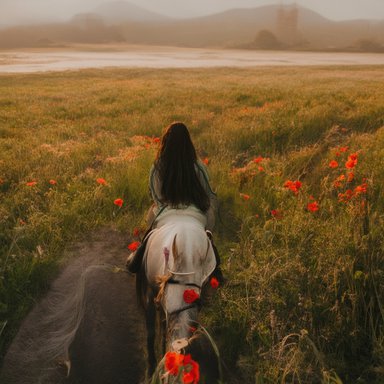}  & \cfgcell{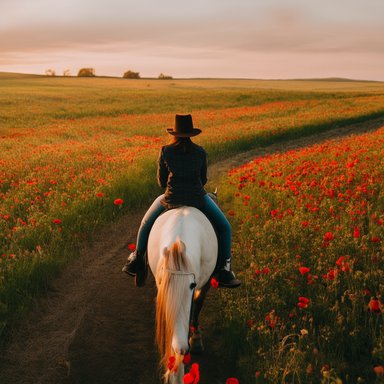}  & \cfgcell{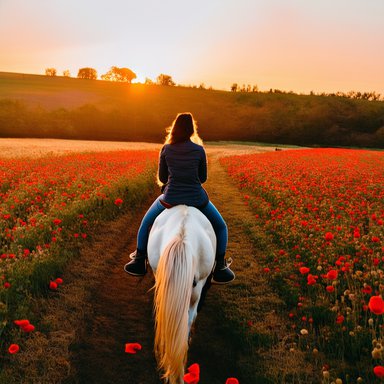}  & \cfgcell{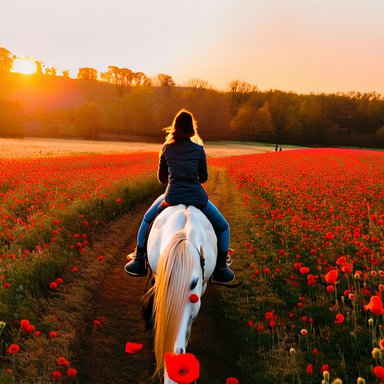} & \cfgspacer \\[-8pt]
\cfgrowlab{SAG}  & \cfgcell{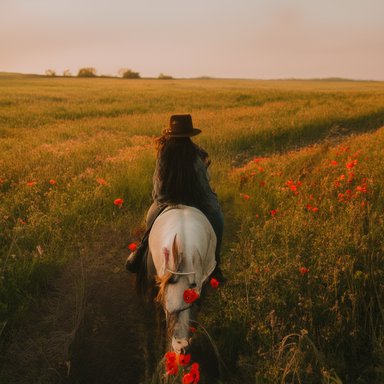}  & \cfgcell{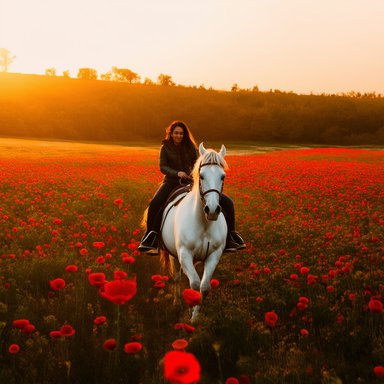}  & \cfgcell{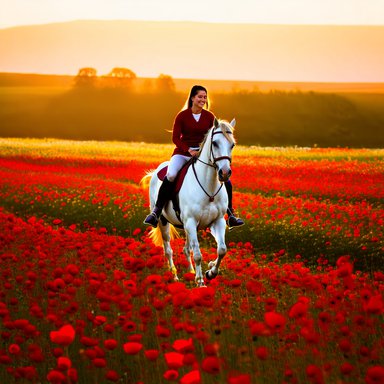}  & \cfgcell{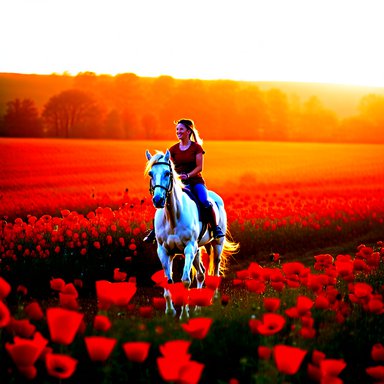} & \cfgspacer \\[-8pt]
\cfgrowlab{OSEG} & \cfgcell{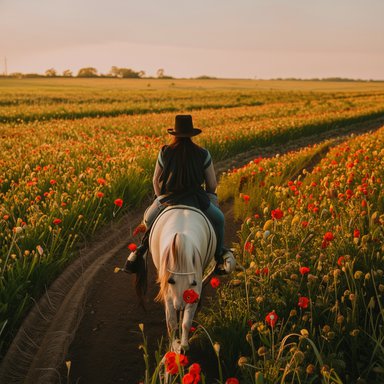} & \cfgcell{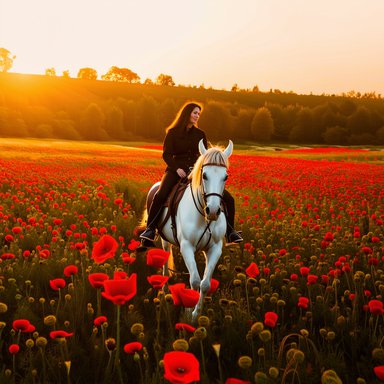} & \cfgcell{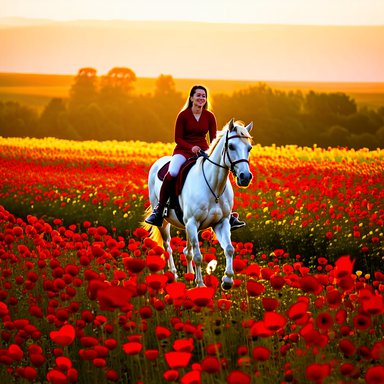} & \cfgcell{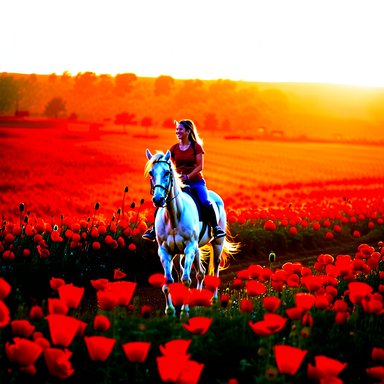} & \cfgspacer \\
\end{tabular}
\caption{Guidance-scale sweep for the white-horse prompt.}
\label{fig:cfgsweep-horse}
\end{figure}

\begin{figure}[!htbp]\centering
\captionsetup{font=large,labelfont=bf,justification=centering,singlelinecheck=false}
\setlength{\tabcolsep}{0pt}\renewcommand{\arraystretch}{0}
\begin{tabular}{@{}c@{}cccc@{}c@{}}
 & \cfgcollab{1.5} & \cfgcollab{5} & \cfgcollab{12} & \cfgcollab{18} & \cfgspacer \\
\cfgrowlab{CFG}  & \cfgcell{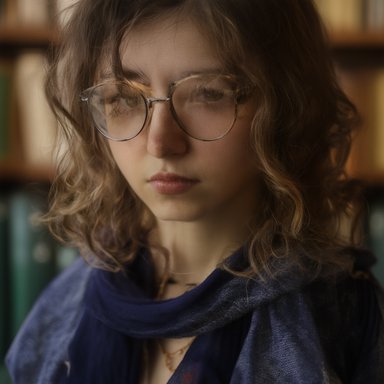}  & \cfgcell{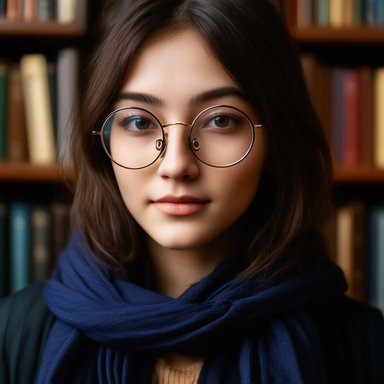}  & \cfgcell{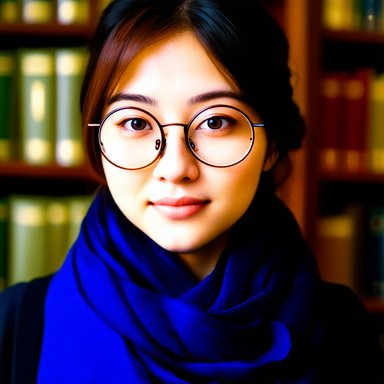}  & \cfgcell{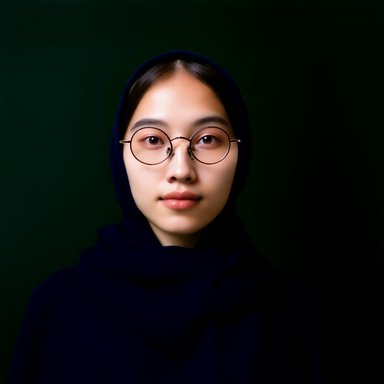} & \cfgspacer \\[-8pt]
\cfgrowlab{APG}  & \cfgcell{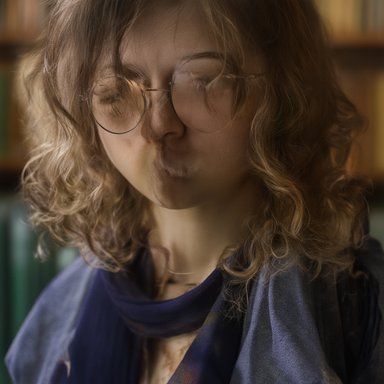}  & \cfgcell{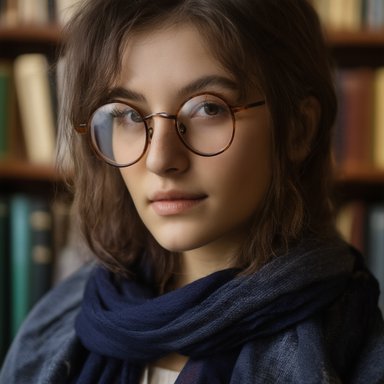}  & \cfgcell{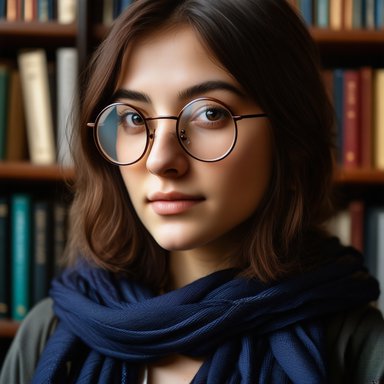}  & \cfgcell{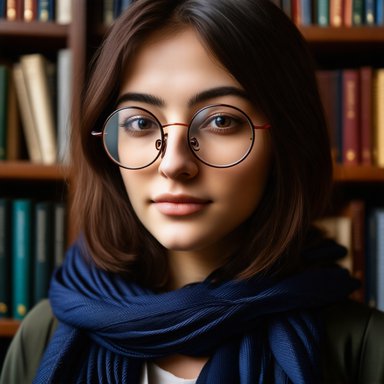} & \cfgspacer \\[-8pt]
\cfgrowlab{SAG}  & \cfgcell{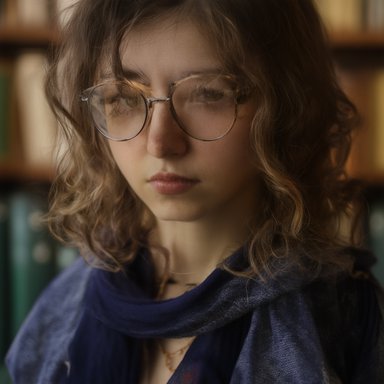}  & \cfgcell{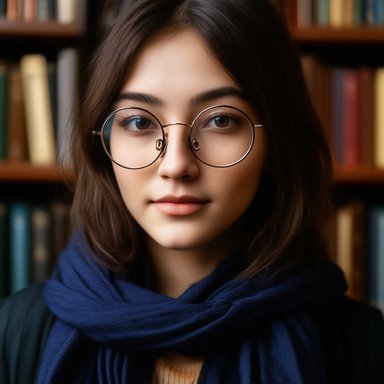}  & \cfgcell{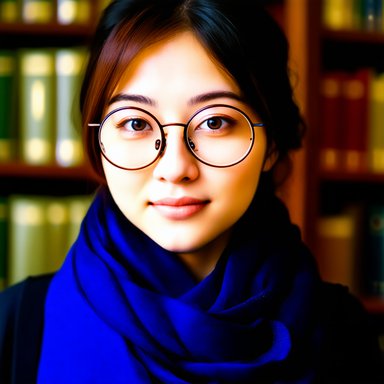}  & \cfgcell{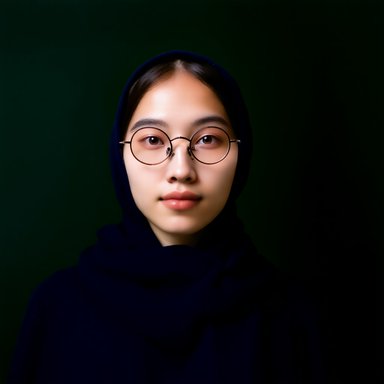} & \cfgspacer \\[-8pt]
\cfgrowlab{OSEG} & \cfgcell{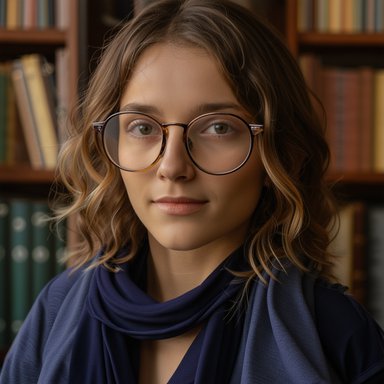} & \cfgcell{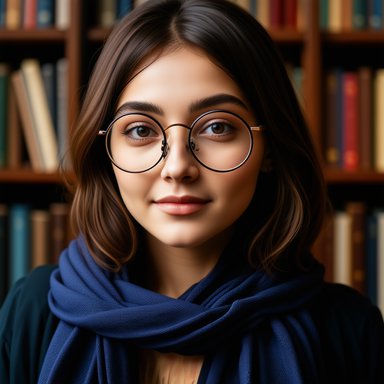} & \cfgcell{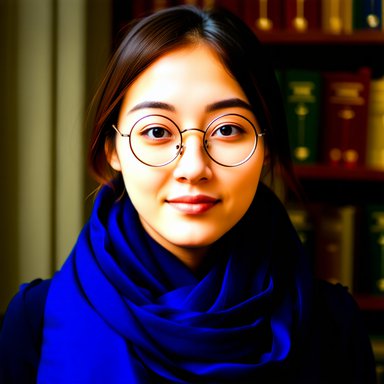} & \cfgcell{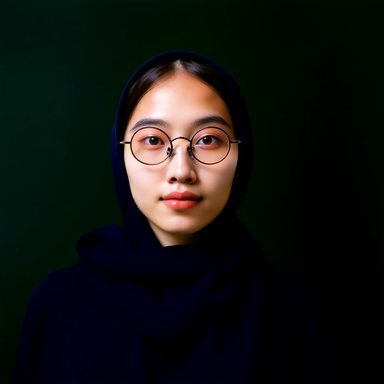} & \cfgspacer \\
\end{tabular}
\caption{Guidance-scale sweep for the portrait prompt.}
\label{fig:cfgsweep-portrait}
\end{figure}

\begin{figure}[!htbp]\centering
\captionsetup{font=large,labelfont=bf,justification=centering,singlelinecheck=false}
\setlength{\tabcolsep}{0pt}\renewcommand{\arraystretch}{0}
\begin{tabular}{@{}c@{}cccc@{}c@{}}
 & \cfgcollab{1.5} & \cfgcollab{5} & \cfgcollab{12} & \cfgcollab{18} & \cfgspacer \\
\cfgrowlab{CFG}  & \cfgcell{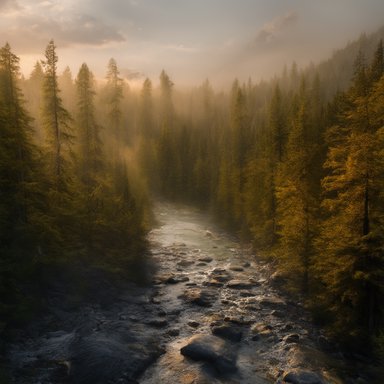}  & \cfgcell{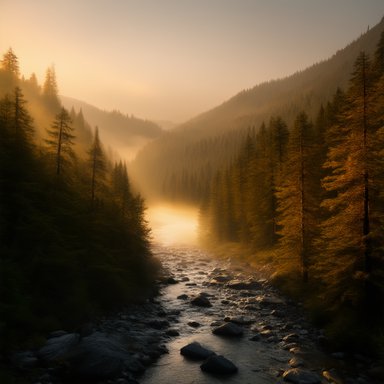}  & \cfgcell{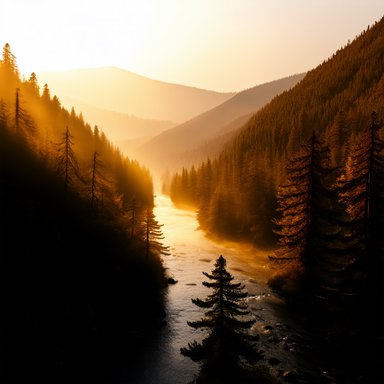}  & \cfgcell{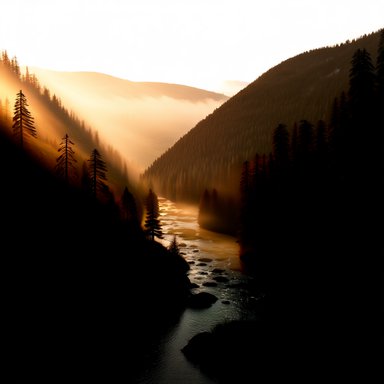} & \cfgspacer \\[-8pt]
\cfgrowlab{APG}  & \cfgcell{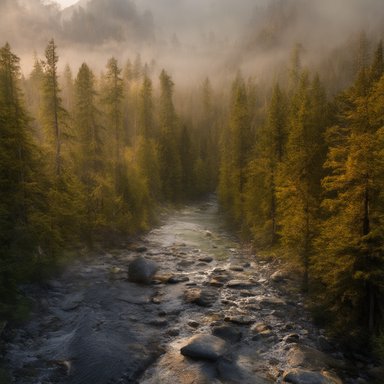}  & \cfgcell{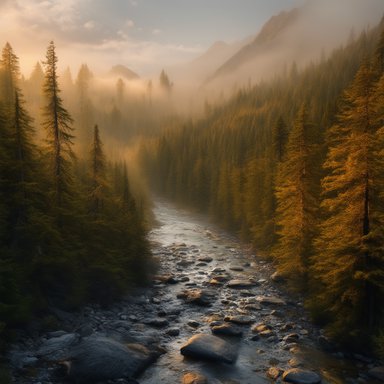}  & \cfgcell{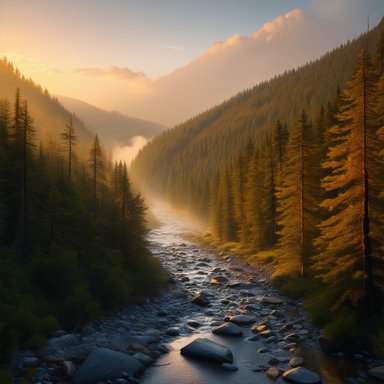}  & \cfgcell{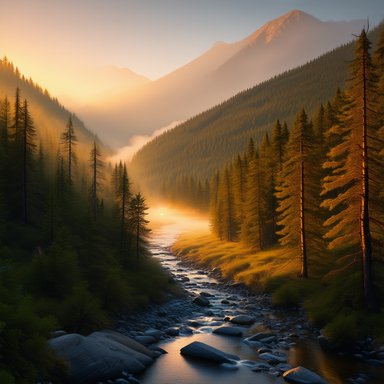} & \cfgspacer \\[-8pt]
\cfgrowlab{SAG}  & \cfgcell{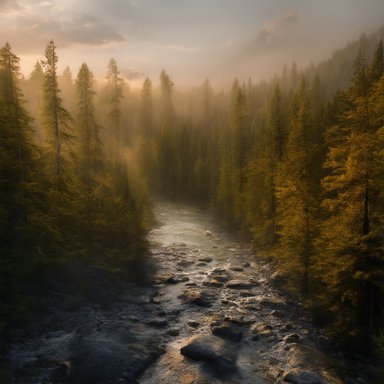}  & \cfgcell{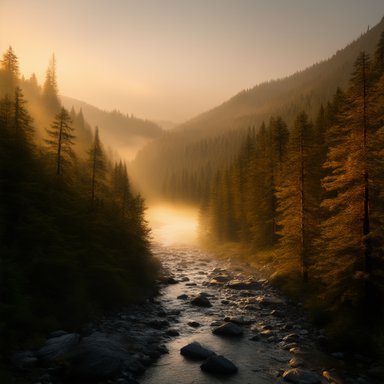}  & \cfgcell{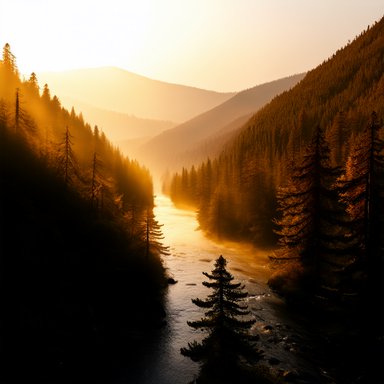}  & \cfgcell{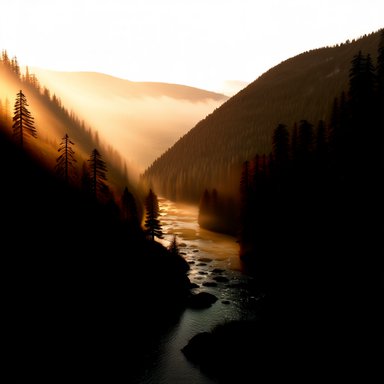} & \cfgspacer \\[-8pt]
\cfgrowlab{OSEG} & \cfgcell{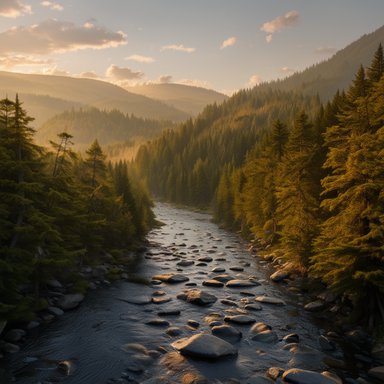} & \cfgcell{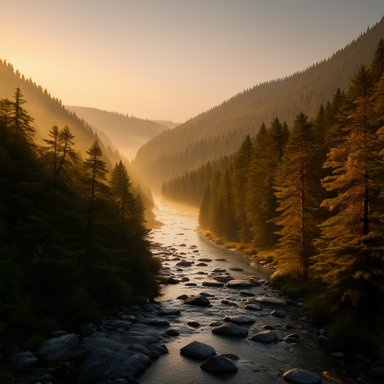} & \cfgcell{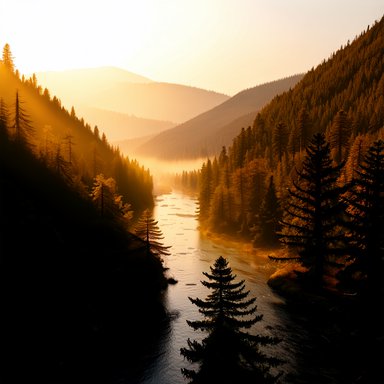} & \cfgcell{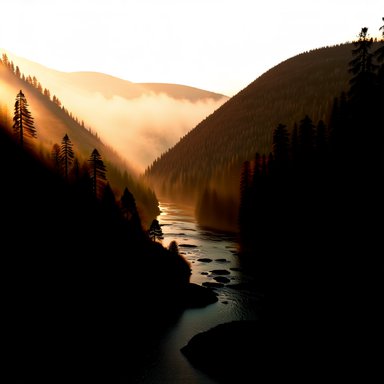} & \cfgspacer \\
\end{tabular}
\caption{Guidance-scale sweep for the mountain-valley prompt.}
\label{fig:cfgsweep-valley}
\end{figure}

\FloatBarrier

\end{document}